**Accepted Version**

# Event-based Civil Infrastructure Visual Defect Detection: ev-CIVIL Dataset and Benchmark

**Authors:**

Udayanga G.W.K.N. Gamage, Xuanni Huo, Luca Zanatta,
T Delbruck, Cesar Cadena, Matteo Fumagalli, Silvia Tolu

# Event-based Civil Infrastructure Visual Defect Detection: ev-CIVIL Dataset and Benchmark

**Udayanga G.W.K.N. Gamage[1], Xuanni Huo[1], Luca Zanatta[2], T Delbruck[3], Cesar Cadena[4], Matteo Fumagalli[1], Silvia Tolu[1]**

**Abstract**

Small Unmanned Aerial Vehicle (UAV)-based visual inspections are a more efficient alternative to manual methods for examining civil structural defects, offering safe access to hazardous areas and significant cost savings by reducing labor requirements. However, traditional frame-based cameras, widely used in UAV-based inspections, often struggle to capture defects under low or dynamic lighting conditions. In contrast, Dynamic Vision Sensors (DVS), or event-based cameras, excel in such scenarios by minimizing motion blur, enhancing power efficiency, and maintaining high-quality imaging across diverse lighting conditions without saturation or information loss. Despite these advantages, existing research lacks studies exploring the feasibility of using DVS for detecting civil structural defects. Moreover, there is no dedicated event-based dataset tailored for this purpose. Addressing this gap, this study introduces the first event-based civil infrastructure defect detection dataset, capturing defective surfaces as a spatio-temporal event stream using DVS. In addition to event-based data, the dataset includes grayscale intensity image frames captured simultaneously using an Active Pixel Sensor (APS). Both data types were collected using the DAVIS346 camera, which integrates DVS and APS sensors. The dataset focuses on two types of defects: cracks and spalling, and includes data from both field and laboratory environments. The field dataset comprises 318 recording sequences, documenting 458 distinct cracks and 121 distinct spalling instances. The laboratory dataset includes 362 recording sequences, covering 220 distinct cracks and 308 spalling instances. We evaluated the dataset using four real-time object detection models. The spatio-temporal event streams from DVS cameras were converted into pseudo-images using encoding methods such as 2D, 2-channel event histograms and fed to these models. The results demonstrate the applicability of DVS cameras for robust detection of civil infrastructure defects under challenging lighting conditions.

*The dataset can be downloaded at https://doi.org/10.11583/DTU.25924228*

*The code together with dataset visualization examples are available at https://github.com/gwgknudayanga/evCIVIL*

**Keywords**

Event-based vision, civil structural health monitoring, defect detection, crack, spalling, DVS, dataset, YOLOv6, SSD, 2D event histograms

## Introduction

Regular civil structural inspection and maintenance are crucial for ensuring safety, mitigating disasters, and safeguarding public welfare. Although 99% of current inspections are manually performed during daytime, recent advancements[1] and ongoing research efforts[76] are increasingly leveraging Unmanned Aerial Vehicles (UAVs) to fully or semi-automate these inspections, thereby improving efficiency, enhancing data reliability, and increasing worker safety while reducing both time and operational costs. However, small UAVs face operational limitations, particularly in battery life and the need for additional controlled lighting in low-visibility conditions (e.g., mines or tunnels, indoor facilities, power plants, or urban structures during night or overcast weather). These limitations hinder the ability to maximize inspection coverage. Traditional solutions, such as frame-based cameras, often fall short due to their low sensitivity and susceptibility to motion blur, making them less effective for deployment on small UAVs. While equipping UAVs with explicit lighting sources, such as strobe lights[2] or spotlights[3], can address low-light conditions, these solutions increase energy consumption, reduce mission duration, and require more missions to inspect a given area. Infrared cameras, widely used in the surveillance industry, offer another solution for low-light condition inspections[4]. However, their low detector sensitivity necessitates longer exposure times, which often results in motion blur when used on aerial drones. In contrast, Dynamic Vision Sensors

[1]Department of Electrical and Photonics Engineering, Technical University of Denmark, Denmark, {kniud, mafum, stolu}@dtu.dk, {huoxuanni}@gmail.com
[2]DEI Department, University of Bologna, Bologna {luca.zanatta3}@unibo.it
[3]Inst. of Neuroinformatics, UZH-ETH Zurich, {todelbru@ethz.ch}
[4]Autonomous Systems Lab, ETH Zurich, {cesarc}@ethz.ch

**Corresponding author:**
Udayanga G.W.K.N. Gamage, Dept. of Electrical Engineering and Photonics, Technical University of Denmark(DTU),
Silvia Tolu, Dept. of Electrical Engineering and Photonics, Technical University of Denmark(DTU)
Email: kniud@dtu.dk, stolu@dtu.dk, gwgknudayanga@gmail.com

(DVS), offer a compelling alternative due to their high dynamic range. They can operate effectively across a wide range of lighting conditions, capturing detailed information in scenarios with extreme brightness, dynamic lighting (where parts of the scene are highly illuminated while others remain dim), or low-light environmentssettings where frame-based cameras typically fail to deliver sufficient detail. Additionally, DVS reduce motion blur compared to both frame-based and infrared cameras in such conditions[5].

Moreover, DVSs offer a significant advantage in power efficiency, consuming approximately 0.1 W [5]. The Basler acA4112-20um monochrome camera, employed as an inspection sensor for railway tunnel inspections, requires 3 W of power[7], which is roughly 30 times higher than the power consumption of DVSs [6].

In surface inspections, where missions often involve uniform surfaces, frame-based cameras capture frames at fixed intervals, recording all intensity information regardless of changes. In contrast, DVS cameras capture only intensity changes as spikes or events, representing increments or decrements in intensity, which helps reduce transmission overhead. Since DVS sensors emit events, we use the terms '***DVS data***' and '***event-based data***' interchangeably throughout this paper, as both refer to the same type of data from DVS.

With additional advantages, such as their ability to operate effectively under dynamic lighting, reduced motion blur, and low power consumption, DVSs are a superior alternative for civil structural inspections. They are particularly suited for challenging lighting conditions, including low-light environments and dynamic scenarios, where energy efficiency is crucial due to limited power budgets. By consuming less power, DVSs extend UAV battery life and mission durations. Furthermore, DVSs inherently require relative motion to detect intensity changes, a condition naturally fulfilled in UAV-based inspections where the UAV moves relative to the structure. This alignment enhances the practicality and effectiveness of using DVSs for UAV-based structural inspections.

As there are no existing studies on event-based defect detection in the current literature, this work aims to address this gap with the following contributions:

- Introduction of the first event-based civil structural defect detection dataset (ev-CIVIL), capturing defective surfaces as a spatio-temporal event stream using a DVS camera. The ev-CIVIL dataset also includes simultaneously captured grayscale intensity image frames using an Active Pixel Sensor (APS), providing a comprehensive dataset for benchmarking.
- Dataset evaluation of defect detection and classification performance using state-of-the-art (SOTA) real-time object detection and classification models.

In our study, the selected object detection models for analyzing the dataset include YOLOv6m, YOLOv6lite-s, SSD300 with ResNet50 backbone, and SSD300 with MobileNetv2 backbone. Among these, YOLOv6lite-s and SSD300 with MobileNetv2 are lightweight models optimized for mobile devices. For classification tasks, we evaluate ResNet34, VGG16, EfficientNet-b0, and MobileNetv2 models, with EfficientNet-b0 and MobileNetv2 being lightweight models specifically designed for mobile devices.

The defect detection and classification results from our comprehensive evaluations demonstrate the applicability of the ev-CIVIL dataset in showcasing the potential of DVSs for civil structural inspections under a wide range of lighting conditions.

The paper is organized as follows: the *"Background"* section provides context and reviews related work, while the *"ev-CIVIL Dataset Description"* section details the ev-CIVIL dataset's composition. The *"Methods"* section outlines the data collection process, preprocessing techniques, deep learning models used for evaluation, and the metrics applied. The *"Experimental Results"* section presents the experiments and their outcomes. The *"Discussion"* section explores the limitations of current approaches, proposes future research directions, and emphasizes the potential of DVSs in civil infrastructure inspection and maintenance. Finally, the *"Conclusion"* summarizes the key contributions and findings of the study.

The overall workflow of this study is summarized in Fig. 1. It encompasses data collection, preprocessing, and labeling to create the ev-CIVIL dataset, extraction of samples from sequences to form the training, validation, and test sets, encoding of event-based data as 2D two-channel histograms, preparation of real-time object detection models, and benchmarking of both event-based and image-based detection performance.

## Background

### *Deep Learning for Visual Civil Infrastructure Inspections*

Visual inspection is fundamental to assessing the condition of civil infrastructures. However, conventional manual methods are labor-intensive, prone to human error, and inherently subjective[8,9]. In contrast, RGB cameras offer a cost-effective and practical alternative due to their ease of deployment and ability to capture detailed visual information[10]. Nevertheless, the efectiveness of this approach depends heavily on the use of advanced image analysis methods. A comprehensive review[11] explores the application of traditional machine learning and computer vision techniques, such as edge detection[12] and Support Vector Machines(SVM)[13], alongside Deep Learning (DL) architectures such as Convolution Neural Networks(CNNs)[14] and Fully Connected Neural Networks(FCN)[15] for defect detection and classification in civil infrastructures. The review highlights the effectiveness of these techniques in detecting and classifying defects such as cracks, spalling in concrete and pavements, and rust in steel structures. Compared to traditional machine learning, the advent of DL has introduced powerful tools for automating defect detection in civil structures. In particular, DL-based object detection frameworks such as You Only Look Once(YOLO)[16] and Single Shot Detector(SSD)[17], along with object classification architectures such as VGG16[18] and ResNet[19], have proven highly effective in this domain[20 21].

While most detection models are designed to identify known defects they were trained on, they often encounter unseen defects (defect types that are not exposed to the model during training) when deployed in real-world

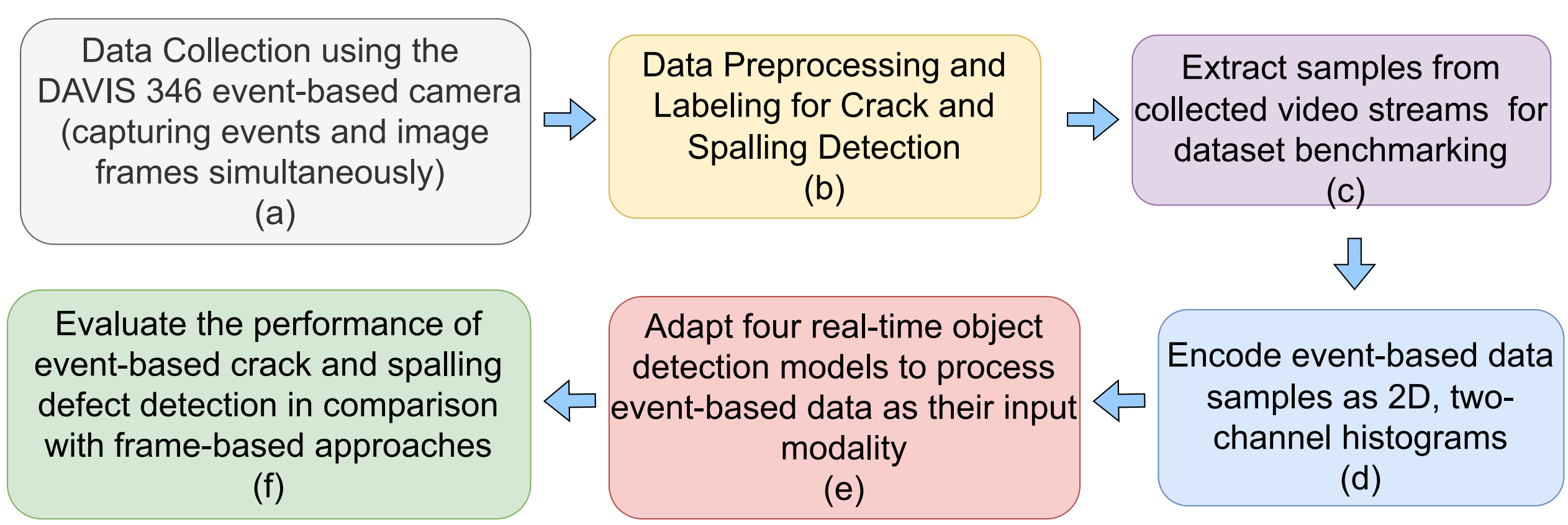

**Figure 1.** Graphical overview of the workflow in our study: (a) both events and grayscale images were captured in the field and in the laboratory to construct the ev-CIVIL dataset (b) illumination enhancement was applied to grayscale frames, event samples with unsynchronized timestamps were removed, and the dataset was annotated for crack and spalling defect detection (c) samples were extracted from the collected video sequences to form the training, validation, and two test sets (d) raw events were encoded as pseudo-images using 2D two-channel histogram representation (e) DataLoaders were implemented for four real-time detection models designed to process event-based data: two YOLOv6 variants (YOLOv6m and lite version) and two SSD variants (ResNet50 and MobileNetv2 based) (f) Benchmark the performance of event-based defect detection against frame-based detection

scenarios. In such cases, these novel defects are typically misclassified as one of the known defect types. In[75], the authors propose a methodology to identify these unseen defects as a separate ”unknown defect” category rather than misclassifying them into known defect types.

All these deep learning models excel in uncovering complex patterns within extensive datasets of labeled images, effectively identifying structural issues such as cracks and spalling. By analyzing footage captured by RGB cameras, DNN models surpass manual inspection methods in several critical aspects. First, they significantly reduce the reliance on human inspectors, thereby enhancing both the efficiency and consistency of defect detection. This is particularly important in scenarios where manual inspection is tedIoUs and prone to oversight. Second, their ability to identify subtle anomalies that might elude human observation improves the accuracy of inspections. This capability is crucial for the early detection of potential structural failures, ensuring timely maintenance and repair. Lastly, the adaptive nature of these networks, which learn and improve with each additional dataset, results in a progressively more accurate and reliable system for inspecting vital civil infrastructure. While prevIoUs studies focused on visual defect classification and detection using images from traditional intensity-frame cameras, the authors in[22] explored civil structural defect classification using event-based data. However, they relied on synthetic event-based data from DVS simulators rather than real data captured by DVS cameras.

## Dynamic Vision Sensors (DVS)

Dynamic Vision Sensors (DVS), also known as event-based cameras, operate asynchronously by detecting changes in brightness at the pixel level. This approach contrasts with traditional cameras, which capture entire frames at fixed intervals. These changes in brightness arise due to relative motion between the camera and the scene, which can result from a moving scene, a moving camera, or a combination of both. When the brightness at a pixel changes beyond a predefined threshold, the event camera instantly records the change, capturing the location and precise timing. This data is logged as either a brightness increment or decrement event[23].

The DVS hardware encodes changes in brightness on a logarithmic scale rather than a linear scale, resulting in a high dynamic range. This characteristic allows it to effectively capture a wide range of light intensities and represent both dim and bright regions in the scene without saturating or losing detail[23]. Due to their event-based operation, DVSs are less susceptible to motion blur compared to frame-based cameras. They can accurately capture fast-moving objects without significant degradation in image quality[5]. Additionally, event-based cameras have minimal latency since they only capture events triggered by changes in brightness, rather than processing entire frames making them ideal for real-time applications where timely response is critical. Furthermore, DVSs consume less power compared to traditional cameras because they only activate when there are changes in the scene. This makes them suitable for battery-powered devices and applications where energy efficiency is important[23].

Nowadays, certain event cameras are engineered not only to capture events but also to record both events and frames simultaneously. For instance, the DAVIS346 camera, shown in Fig. 2, developed by iniVation[25], incorporates both a Dynamic Vision Sensor (DVS) and an Active Pixel Sensor (APS). This allows it to capture grayscale image frames alongside asynchronous events associated with changes in brightness. In addition to iniVation, Prophesee is another prominent manufacturer of dynamic vision sensors.

As shown in Fig. 2, the events generated by DVSs form a spatio-temporal event stream, where the x and y axes represent spatial coordinates in the pixel frame, and the time axis represents the events timing. These spatio-temporal event volumes can be organized into 2D event histograms, as visualized in Fig. 2, and subsequently fed to learning algorithms for tasks such as detection and classification. The process of forming 2D event histograms is described in detail in the Methods section.

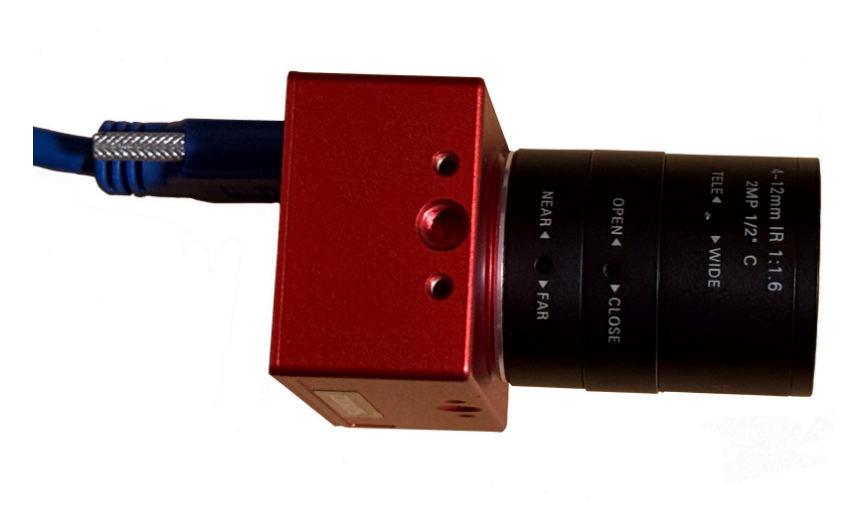

(a) DAVIS346 camera

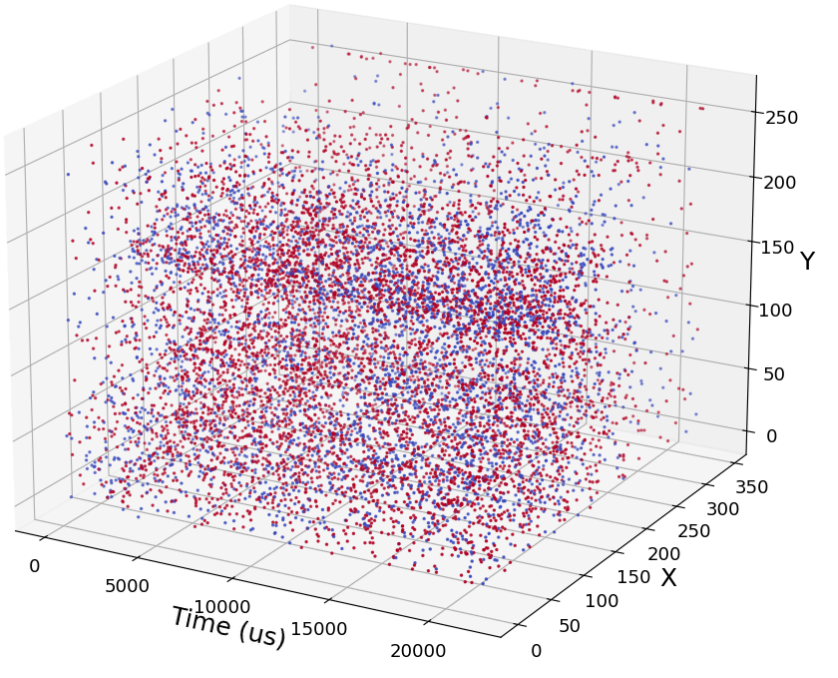

(b) spatio-temporal event volume

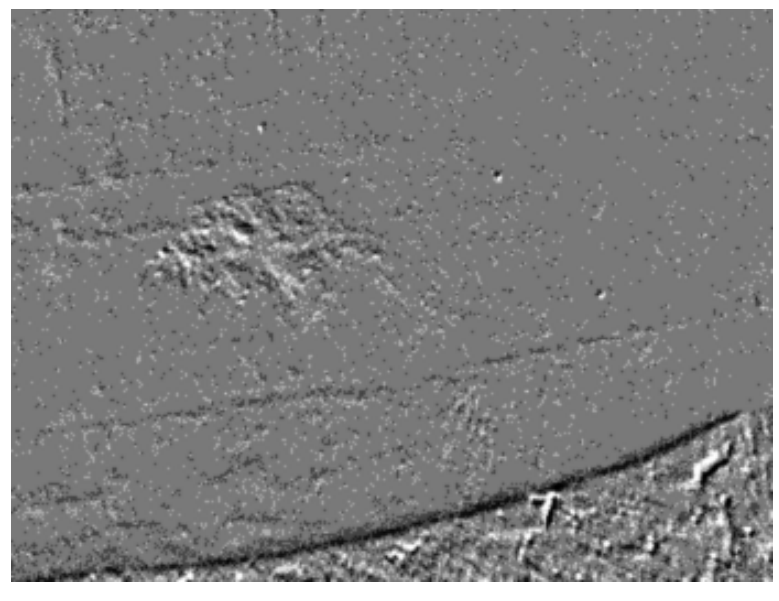

(c) 2D event histogram visualization

**Figure 2.** DAVIS346 event volume formation: Spatio-temporal event volumes (b) captured by the DVS camera (a) are encoded as 2D, 2-channel histograms (c) using Algorithm 1. For (c), events occurring over 15ms in the 2D pixel space (346260) of the DAVIS346 camera are used.

## *Event-based Datasets for Object Detection and Classification*

Most of the recently developed DVS datasets for object detection and classification have focused on automotive applications. Table 1 provides a comprehensive summary of event-based detection datasets and introduces the algorithms used to evaluate them, along with the ***'ev-CIVIL'*** dataset introduced in this paper.

The Prophesee GEN1 automotive dataset, introduced in [26], consists of 39 hours of recordings aimed at detecting cars and pedestrians. In [72], the GEN1 dataset was assessed using the SSD architecture with various backbones, such as VGG and MobileNet, achieving a maximum $mAP_{0.5:0.95}$ value of 0.19. In [73], the same dataset was evaluated using YOLOv6, yielding a $mAP_{0.5:0.95}$ value of 0.5.

In [27], the Prophesee 1 MPixel dataset is introduced, featuring high-resolution recordings totaling 15.65 hours. This dataset is designed for detecting pedestrians and seven distinct vehicle types in driving scenarios. It was evaluated using an architecture called RED [27], which combines CNN and Recurrent Neural Network (RNN) blocks, focusing on three object classes and achieving a mAP of 0.4. In [73], the same 1 MPixel dataset was evaluated with YOLOv6, achieving a $mAP_{0.5:0.95}$ detection performance of 0.4.

In [32], the DSEC dataset is introduced, incorporating LiDAR, GPS, and stereo recordings from DVSs and RGB cameras, primarily for autonomous driving applications. This dataset was subsequently extended with enhancements detailed in [28] to support object detection tasks. In [74], the dataset was evaluated using a CNN-based object detector with a ResNet-101 backbone and custom feature fusion, achieving a $mAP_{0.5}$ of 38.38.

In [29], the authors introduced an event-based dataset for vehicle detection (single class). The dataset was evaluated using SSD and YOLOv3 architectures. Instead of the $mAP$ metric, the authors assessed detection performance using two metrics: $DetA$ (detection accuracy) and $LocA$ (localization accuracy). In their evaluation, they achieved a maximum of 52 for $DetA$ and 84 for $LocA$ with YOLOv3.

In [30], the PEDRo dataset is introduced, focusing on person detection within robotics applications. This dataset was evaluated using the YOLOv8x architecture, achieving a $mAP_{0.5:0.95}$ value of 0.58 for person detection.

Additionally, [33] describes an effort where the authors annotated 368 nighttime frames from the MVSEC stereo DVS dataset, which was originally designed for 3D perception. This work let to the creation of the MVSEC-NIGHTL212 dataset. The evaluation of car detection performance on this dataset, including both day and night samples, yielded average precision at 50% intersection over union ($AP_{50}$) values of 0.35 and 0.30 for the day and night test sets, respectively, using the YOLOv3 model.

The Prophesee NCAR dataset, detailed in [31], focuses on car classification and contains 24,029 samples, each lasting 100 ms, totaling approximately 40 minutes of recordings. The dataset was evaluated using both a linear SVM [13] classifier and a deep CNN-based classifier, achieving a classification accuracy of 90%. In addition to automotive datasets, researchers have also developed laboratory-based DVS datasets for classification tasks by displaying static images from sources like the MNIST and Caltech101 datasets on a computer screen and capturing the scene with an event camera during various movements. In [35], three saccadic movements were performed with the event camera aimed at the display, simulating rapid eye movements to capture dynamic changes in the scene. In contrast, [36] explored repeated Closed-Loop Smooth (RCLS) movements of images displayed on a computer monitor. These moving images were captured using a stationary event camera.

In [34], the authors introduced the MED3 DVS dataset, which was created through robot navigation in diverse indoor and outdoor environments, both during the day and at night. While the dataset includes ground truth data for Simultaneous Localization and Mapping (SLAM), it does not provide ground truth annotations for object detection tasks.

A comprehensive comparison of our ev-CIVIL dataset with other mentioned event-based object detection datasets is provided in Table 1 and further detailed in the "ev-CIVIL Dataset Description" section. This analysis highlights the unique features and advantages of the ev-CIVIL dataset in relation to existing datasets, emphasizing its relevance and contributions to advancing event-based data research.

**Table 1.** Comparison of Event-based Datasets for Object Detection

| Attributes | Prophesee GEN1 [26] | Prophesee 1MPixel [27] | DSEC Detection [28] | Event-based vehicle detection and tracking [29] | PEDRo [30] | ev-CIVIL (Ours) |
|---|---|---|---|---|---|---|
| Task | Autonomous object detection | Autonomous object detection and tracking | Autonomous object detection and tracking | Autonomous object detection and tracking | Person detection in robotics | Civil Infrastructure Defect Detection |
| Light Conditions | Day light Night ambient light | Day light Night ambient light | Day light Night ambient light | Day light | Day light Night ambient light | Day light Night ambient light **Laser light at night** |
| # of Sequences | - | - | 60 (in-field) | 31 (in-field) | 119 (in-field) | **318** (in-field) 361 (in-lab) |
| # of Hours | 39 | 15 | - | - | 0.3 | 1.3 (in-field) |
| # of Classes | 2 | 7 | 8 | 1 | 1 | 2 |
| # of bbox Annotations | 255K (GT) | 25M (Automatic) | 390118 (GT) | 9891 (GT) | 43259 (GT) | 62276 (in-field) 138979 (in-lab) |
| Annotation Rate(Hz) | 1 - 4 | 60 | 20 | 24 | 25 | 1 - 30 |
| Spatial Resolution | 304×240 | 1280×720 | 640×480 | 240×180 | 346×260 | 346×260 |
| Deep learning architectures used to evaluate | SSD,YOLOv6 | RED(custom), YOLOv6 | custom CNN with Resnet101 backbone | YOLOv3,SSD | YOLOv8 | YOLOv6, SSD |

## *ev-CIVIL* Dataset Description

***ev-CIVIL*** refers to our dataset, which includes event-based data captured by a DVS sensor with grayscale intensity image frames simultaneously capturedby an APS sensor, both integrated within the DAVIS346 camera. The dataset is specifically designed to address two extensively documented types of civil infrastructure defects: ***cracks*** and ***spalling***. It predominantly conprises field data, showcasing real-world defect scenarios in civil infrastructure, supplemented by laboratory data to enhance its scope. The following sections elaborate on the significance and details of these components, emphasizing their importance in building a robust and versatile dataset for defect detection research and applications.

Our dataset comprises 318 recording sequences collected from the field, showcasing defects in various types of civil structures, including roads, pavements, tunnels, buildings, and walls containing 458 unique crack instances and 121 spalling instances. Examples of these data samples are visualized in Fig. 3a. The figure displays grayscale image frames captured by the APS sensor, denoted as 'fr' in the corresponding columns. Data captured by the DVS sensor are represented as 2D event histograms and are visualized in columns labeled 'ev' within the same figure. The field dataset comprises a total recording time of approximately 80 minutes and includes 62,276 ground truth bounding boxes: 44,880 for cracks and 17,396 for spalling. This extensive collection provides a solid foundation for event-based civil infrastructure defect detection and analysis.

Due to the insufficient number of unique crack and spalling instances in the field data component to effectively train state-of-the-art deep neural networks for object detection with high accuracy, we augmented the ev-CIVIL dataset by introducing an additional component, referred to as the laboratory dataset. Although there are currently no existing event-based datasets specifically for civil infrastructure defect detection, several image-based datasets for crack and spalling defects are available in the literature. Using these image-based datasets, we enhance the diversity and quantity of data in the ev-CIVIL dataset. This augmentation enhances the dataset's suitability for training advanced deep neural networks, enabling improved accuracy in defect detection tasks.

Typically, in visual inspections of civil structures, the images captured by cameras are two-dimensional (2D). "Why not take images from existing datasets, project them onto 2D surfaces, and capture them with a DVS sensor to understand how DVS sensors perceive cracks and spalling?" This idea forms the foundation of our laboratory dataset, as detailed further in the Subsection "Data Collection Approach". In the literature, event-based simulators such as v2e have been used to generate event-based data from images. However, in this study, we employ real DVS hardware, consistent with prevIoUs works such as[35] and[36], as this approach accounts for the noise and hardware considerations inherent to DVS sensors. As summarized in Table 1, the laboratory dataset comprises 361 sequences, documenting 220 distinct cracks and 308 distinct instances of spalling. It includes a total of 138,979 ground truth bounding boxes: 57,671 for cracks and 81,308 for spalling. Selected samples of crack and spalling from the laboratory dataset are shown in Fig. 3b. In these samples, columns denoted as 'fr' represent the grayscale image frames captured by the APS sensor, while the 'ev' columns display the 2D event-histogram visualizations, following the same conversion used for field data visualization.

As highlighted in Table 1, the 'ev-CIVIL' dataset is the first event-based dataset specifically designed to detect civil infrastructure defects. According to our literature survey, it distinguishes itself by offering the largest collection of recorded sequences among existing event-based field datasets.

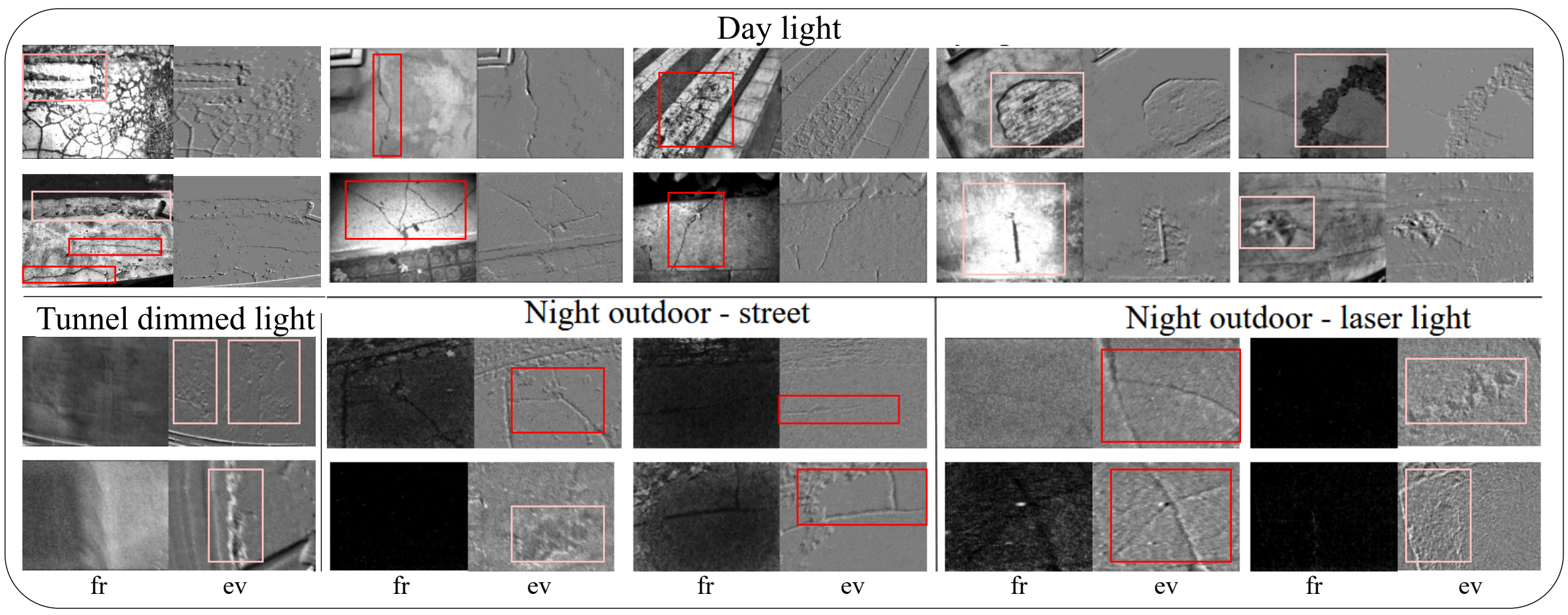

**(a)** Examples from the field data: Crack regions are highlighted with red bounding boxes, while spalling regions are marked with pink bounding boxes. Except for the daylight case, these annotations are shown on the 2D event histograms (ev), because in those scenarios the grayscale image (fr) quality is inferior to that of the 2D event histograms.

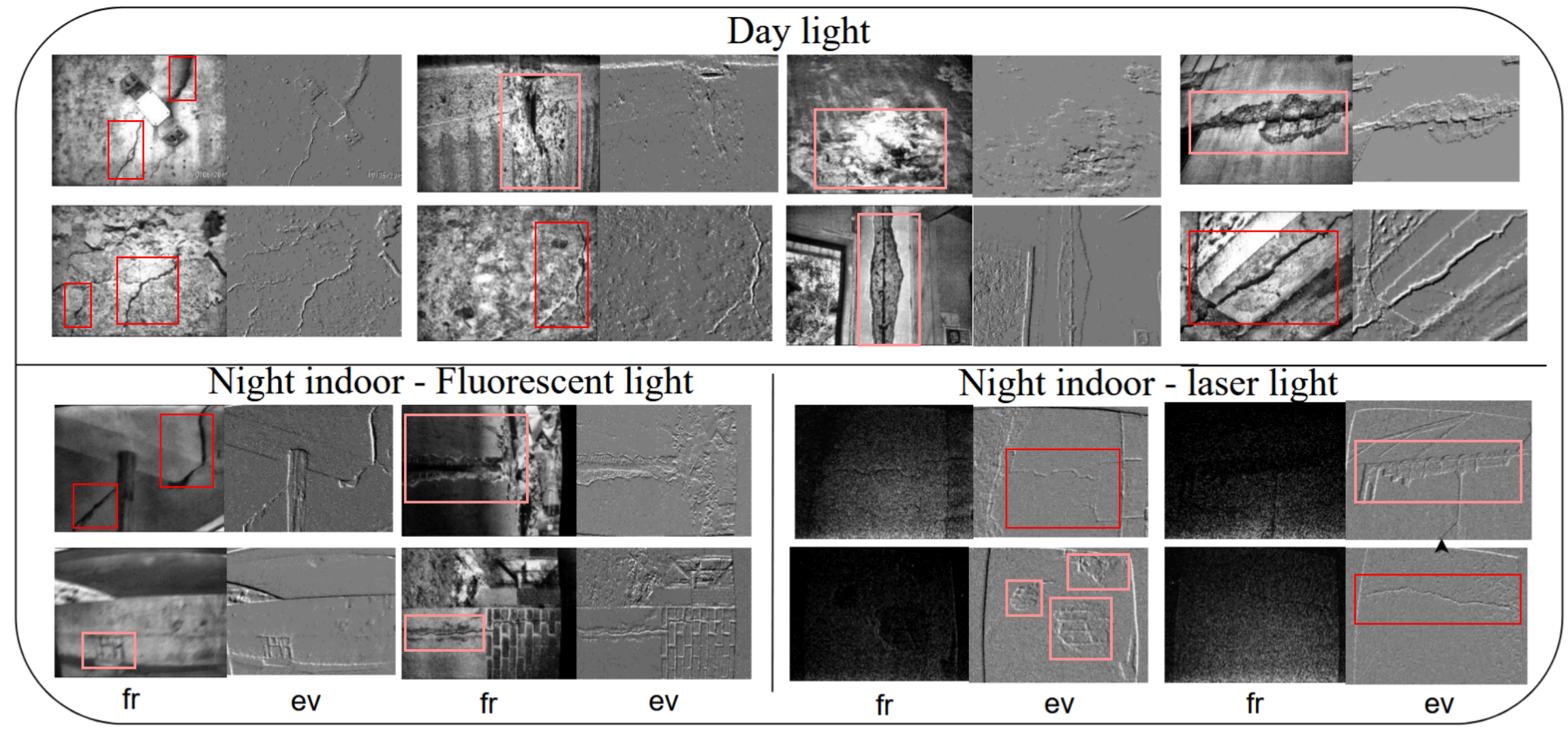

**(b)** Examples from the laboratory data: Cracks are highlighted with red bounding boxes, while spalling regions are marked in pink. When the visual quality of the grayscale image frames (fr) is inferior to that of the 2D event histograms (ev), the bounding box annotations are drawn on the event histograms; otherwise, they are shown on the grayscale images.

**Figure 3.** Field and Laboratory data examples

The dataset captures defects under a diverse range of lighting conditions, including scenarios where an energy-efficient external light source was used to compensate for insufficient ambient lighting. In some cases, the ambient lighting was so limited that neither the APS nor the DVS could capture any meaningful information. To address this, we used the low-power structured infrared laser (IR laser) integrated into the Intel D435 camera[37] as the external illumination source. Details on the integration and use of the Intel D435 in our data collection process are provided in the "Methods" section.

Fig. 4 provides a summary of the number of data samples collected under different lighting conditions during field and laboratory data collection efforts, highlighting the dataset's adaptability to diverse environmental scenarios As illustrated in Fig. 4, the majority of the field data were collected under diverse daylight conditions, including sunny, bright sunlight, cloudy, partly cloudy, and dusk/sunset scenarios. Beyond daytime sequences, the dataset also includes outdoor night sequences, with 19 captured under street lighting and 65 recorded using laser lighting. Additionally, the dataset incorporates a limited number of sequences collected inside a tunnel, where only dimmed tunnel lights were used without any external light sources.

In the laboratory dataset, 146 sequences were collected under fluorescent lighting conditions inside an office, specifically during cloudy weather or at night. In contrast, 145 sequences were recorded when natural daylight was sufficient to illuminate the office. Additionally, laser lighting served as the primary indoor light source in the absence of other lighting, contributing to 70 recorded sequences. This diverse range of lighting conditions in both the field and laboratory datasets underscores the comprehensiveness and versatility of the dataset.

Fig. 5 shows the event rate distribution under different lighting conditions based on the ev-CIVIL event-based dataset. To build the histograms, we used samples from both the training and test sets employed for benchmarking. The highest mean spike rate occurs in data captured under fluorescent lighting, likely because the flickering frequency of fluorescent lights can trigger noisy events in DVS sensors. In data collected under street lighting, three smaller clusters of histograms can be observed, which may correspond to samples captured at varying distances from the street lights. Additionally, some samples collected inside dimly lit tunnels are included in these clusters, although their number is relatively small. The lowest mean event rate is observed in data captured under daylight, which has relatively decent illumination and thus less DVS noise, followed by samples collected under laser lighting.

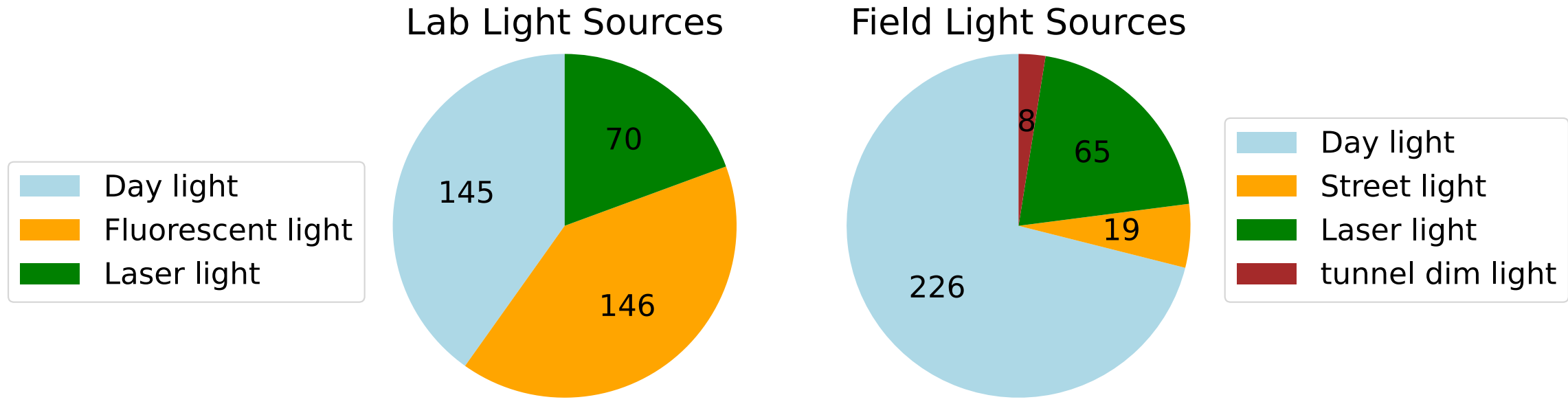

**Figure 4.** Number of sequences collected under different lighting conditions.

### *Dataset Deployment*

Each recording sequence in the dataset primarily consists of three components, as illustrated in Fig. 6a:

- An .h5 file containing event data (e.g., events.h5).
- An .h5 file containing captured grayscale image frames (e.g., frames.h5).
- An .npy file containing bounding box annotations for identifying crack and spalling defects, formatted according to the COCO bounding box convention (e.g., label.npy).

In the events .h5 file, each event is represented by four elements: timestamp, x and y pixel coordinates, and p event polarity.

- Timestamp: indicates the time ($\mu$s) at which the event was triggered and captured.
- x, y: Represent the 2D pixel coordinates of the event within the DAVIS346 cameras pixel array ($346 \times 260$).
- Polarity: Specifies the nature of the event: A polarity value of 1 indicates an increase in perceived pixel intensity. A polarity value of 0 indicates a decrease in perceived pixel intensity.

In the frames .h5 file, each entry contains two elements:

- A grayscale image frame, represented as a 2D array of integer values with dimensions ($346 \times 260$), which encodes the intensity values of each pixel.
- Timestamp indicating when the frame was captured by the camera.

Annotations for recording sequences are provided relative to the timestamp of a specific event and are stored in a .npy file. Each row in the file corresponds to a separate bounding box annotation, identifying either a crack or spalling. For each annotation (row), the columns are structured as follows:

- Column 1: The timestamp of the event to which the bounding box annotation corresponds.
- Column 2: The *class_id*, where 0 represents a bounding box encapsulating a crack, and 1 represents a bounding box encapsulating spalling.
- Columns 3-6: Bounding box information in the 'COCO' annotation format.

In Fig. 6b, an example is presented that illustrates the content of the events.h5, frames.h5, and label.npy files. At a given timestamp, multiple events are triggered from different pixels. By analyzing the timestamps, it is evident that the resolution for capturing events is approximately 1 ms. In contrast, for the image frames in this particular example, the interval between consecutive frames is around 30 ms.

To find the annotation for a specific grayscale image frame with a timestamp $t$ (in milliseconds), it is necessary to identify timestamps in the .npy file that fall within the range $[t - 10, t + 10]$ ms and extract the corresponding annotations. Similarly, for locating annotations corresponding to a 2D event histogram, the minimum and maximum timestamps of all events contributing to the histogram (denoted as $t_{min}$ and $t_{max}$) are identified. Annotations with timestamps within the range $[t_{min}, t_{max}]$ are then extracted.

In most cases, annotations for both events and grayscale images within a recording sequence are stored in the same .npy file. However, during the analysis of recorded data, 30 sequences were identified where the grayscale images and their corresponding events were not properly time-synchronized. For these sequences, separate .npy annotation files were created: one for the grayscale image frames and another for the event data captured by the DVS sensor. These annotations were manually created and independently processed following the procedures explained in the "Data Annotation" subsection of the methods section.

## Methods

This section describes the methods used in both the data collection and dataset benchmarking processes.

### *Data Collection*

For our study, which evaluates event-based defect detection in comparison to grayscale image-based detection, we specifically required a ***hybrid event-based camera*** capable of capturing both events and grayscale image frames simultaneously. To meet this requirement, we selected the DAVIS346 event-based camera from iniVation, which integrates both DVS and APS technologies. At the time of purchase, it was the latest hybrid event-based camera from iniVation[25] that could capture both modalities simultaneously.

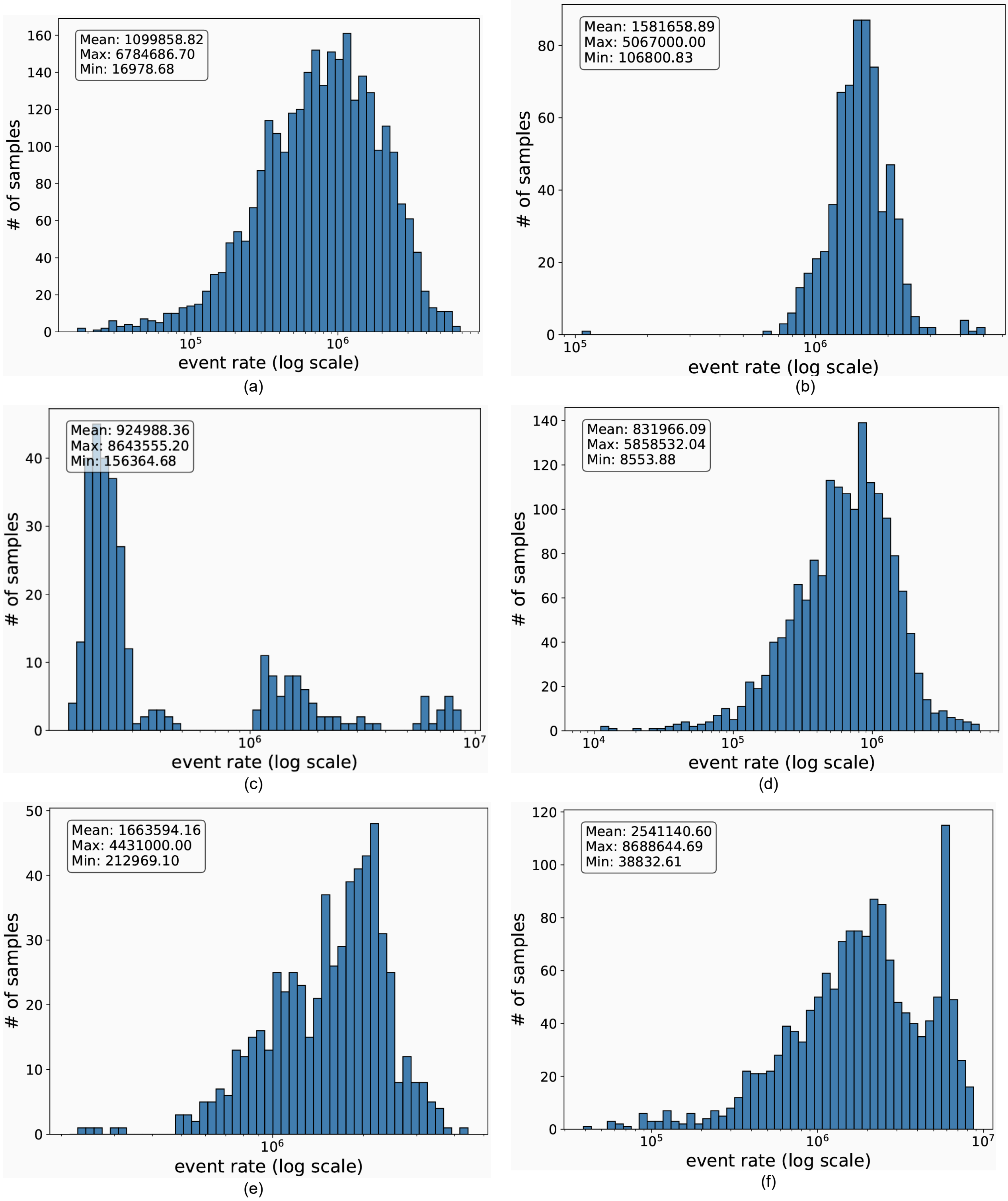

**Figure 5.** Distribution of event rates under different lighting conditions: (a) field-daylight, (b) field-laser light, (c) field-street light, (d) laboratory-daylight, (e) laboratory-laser light, and (f) laboratory-fluorescent light.

*Equipment Setup* Fig. 7a shows the setup used for capturing a defective surface in the field. During a recording sequence, we moved the DAVIS346 camera along a trajectory resembling a “Z-shaped camera trajectory with smooth, flowing bends” as shown in Fig. 8a. The motion began with a near-horizontal movement, approximately parallel to the defect, at a distance of 60–70 cm from the defect surface. The camera then moved closer to the defect

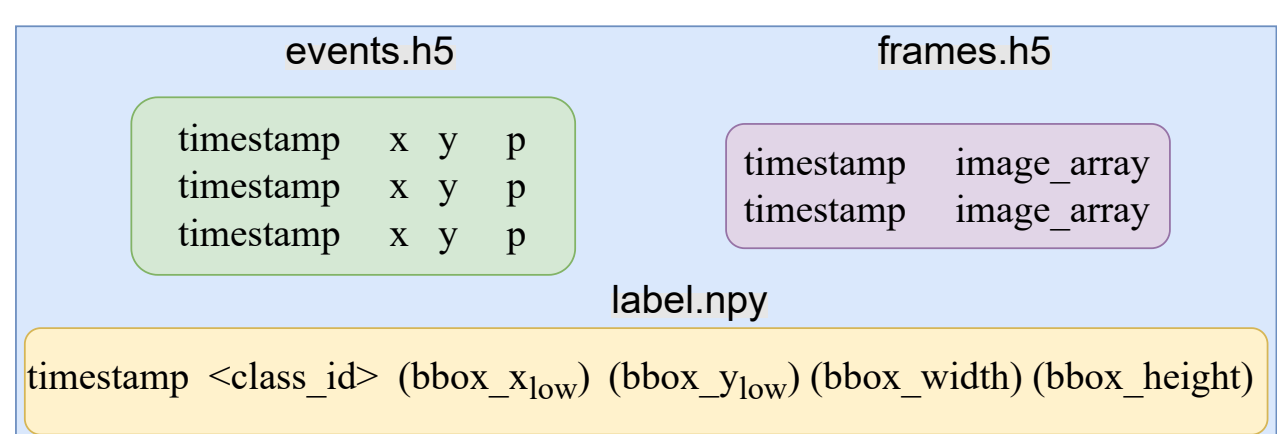

**(a)** Template outlining the composition of files

**(b)** Specific example illustrating the content of files

**Figure 6.** Structure of a recording sequence in the ev-CIVIL Dataset: each event is characterized by a timestamp in $\mu s$, x and y are pixel coordinates within the 346x260 DAVIS346 spatial resolution, and a polarity value (p). The polarity p indicates the type of event: 1 for an increase in pixel intensity and 0 for a decrease.

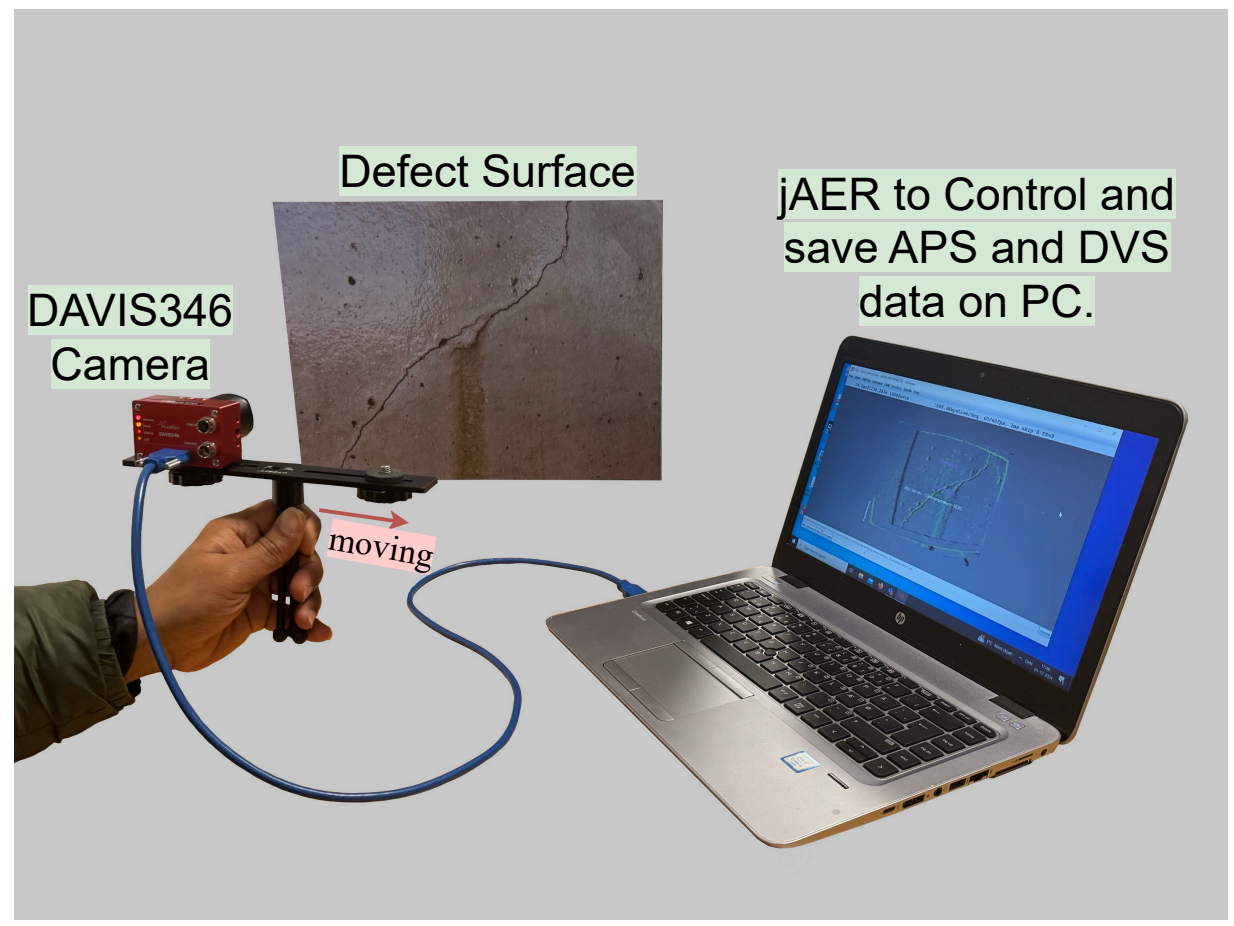

**(a)** Field data collection setup

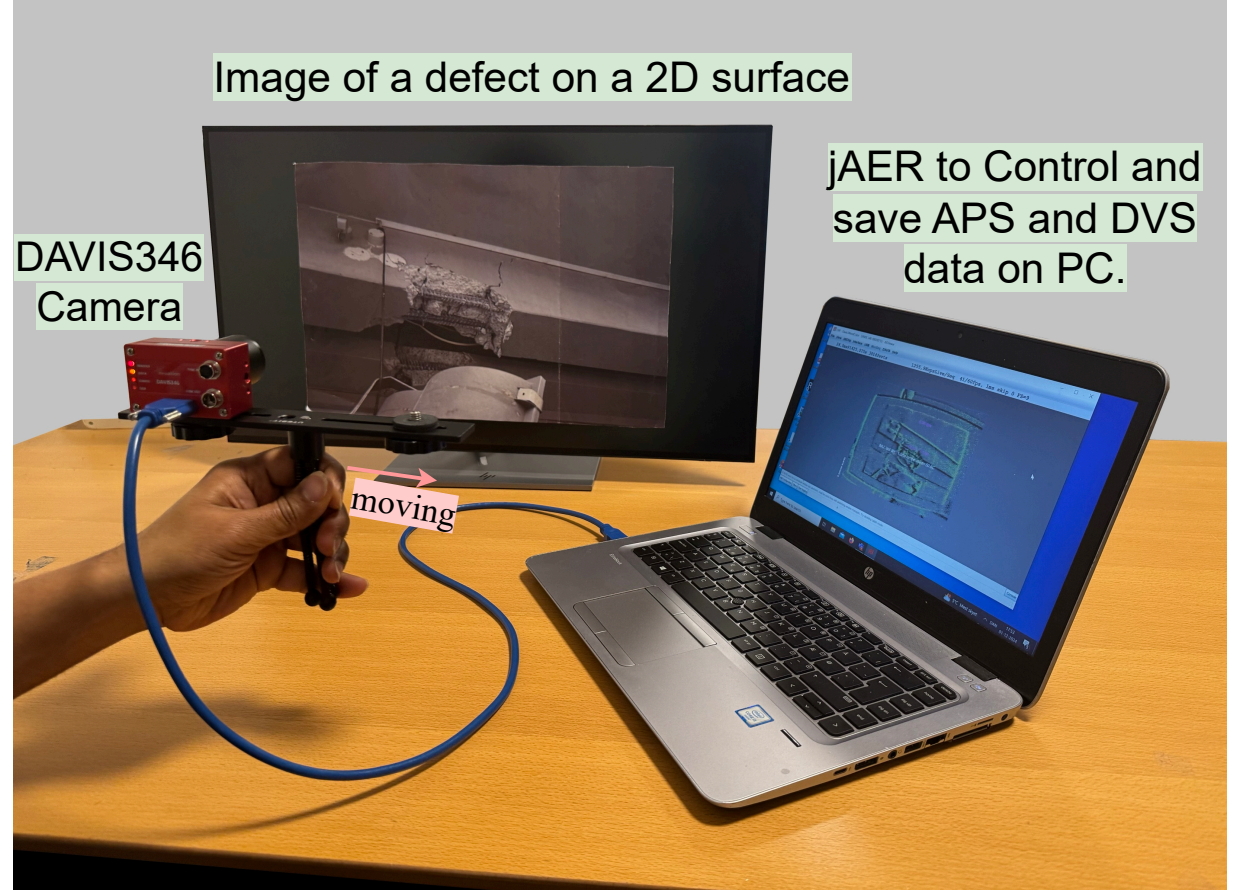

**(b)** Laboratory data collection setup and continued with another horizontal motion, maintaining a distance of around 20–30 cm cm from the defect surface. All movements were conducted handheld.

**Figure 7.** Data Collection Setup

The trajectory shown in Fig. 8a was drawn using accelerometer and gyro data obtained from the IMU sensor integrated into the DAVIS346 camera. Based on the same IMU data, the velocity magnitude of the camera is illustrated in Fig. 8b. The average velocity magnitude of the motion was approximately 1.1m/s. By selecting 50 IMU recordings corresponding to 50 field sequences, we observed that the average velocities of the sequences ranged from approximately 0.5–1.6m/s. Based on 50 laboratory-based recordings, the average velocities ranged from 0.35–1.2m/s. These values are approximate due to potential errors in the IMU data. Along with average velocity ranges, trajectory types, and the distance range from the defective surface to the camera, Table 2 summarizes other key operational and environmental conditions related to data collection, such as the textures of the defect surfaces, lighting, and environmental conditions. The data collection process spanned several months across different European regions, including Switzerland, Denmark, Spain, and Portugal.

While following the trajectory, the camera captures events and grayscale frames simultaneously and saves them to the Personal Computer (PC). To control the data capture and saving process, either jAER[44] software or ROS-DVS[56] software was used.

Fig. 7b shows the setup used to capture a defective surface in the laboratory. The key difference compared to the field setup is that, in the laboratory, we used high-resolution 2D images of cracks or spalling as the defect surfaces. As introduced in *"ev-CIVIL Dataset Description"* section, we started by printing high-resolution images of selected cracks and spalling from existing frame-based civil structural inspection datasets[41 42 43] onto A3-sized pages. These images were then affixed to a black 2D surface, as shown in Fig. 7b, and captured by moving the camera along the prevIoUsly mentioned trajectories.

Fig. 9 shows a detailed overview of the data collection methodology used to gather data for the ev-CIVIL dataset. The process includes preparing the DAVIS346 camera, simultaneously capturing and saving events and grayscale image frames, and subsequent data labeling. The following subsections provide a detailed explanation of these steps.

*DAVIS346 camera preparation* The camera preparation step includes configuring the necessary settings of the DAVIS346 camera itself and incorporating the external laser light source in scenarios where either APS or DVS fails to capture any meaningful information due to insufficient ambient lighting.

To configure the DAVIS346 for data collection, we adjusted the bias values, optimized focus settings, and performed calibration and distortion correction separately for daytime and nighttime scenarios. For data collection

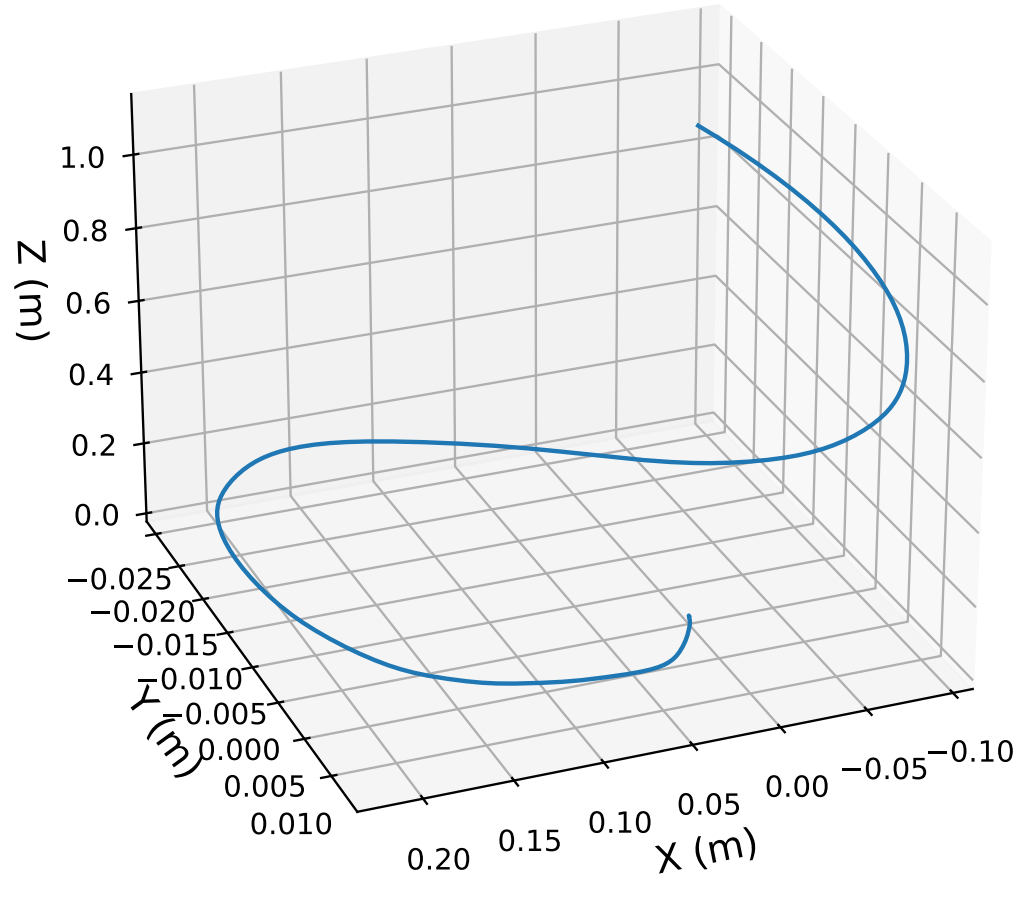

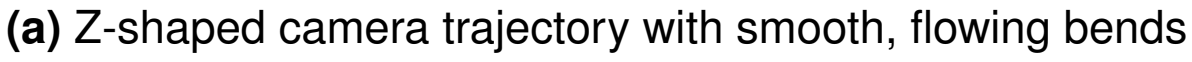

**(a)** Z-shaped camera trajectory with smooth, flowing bends

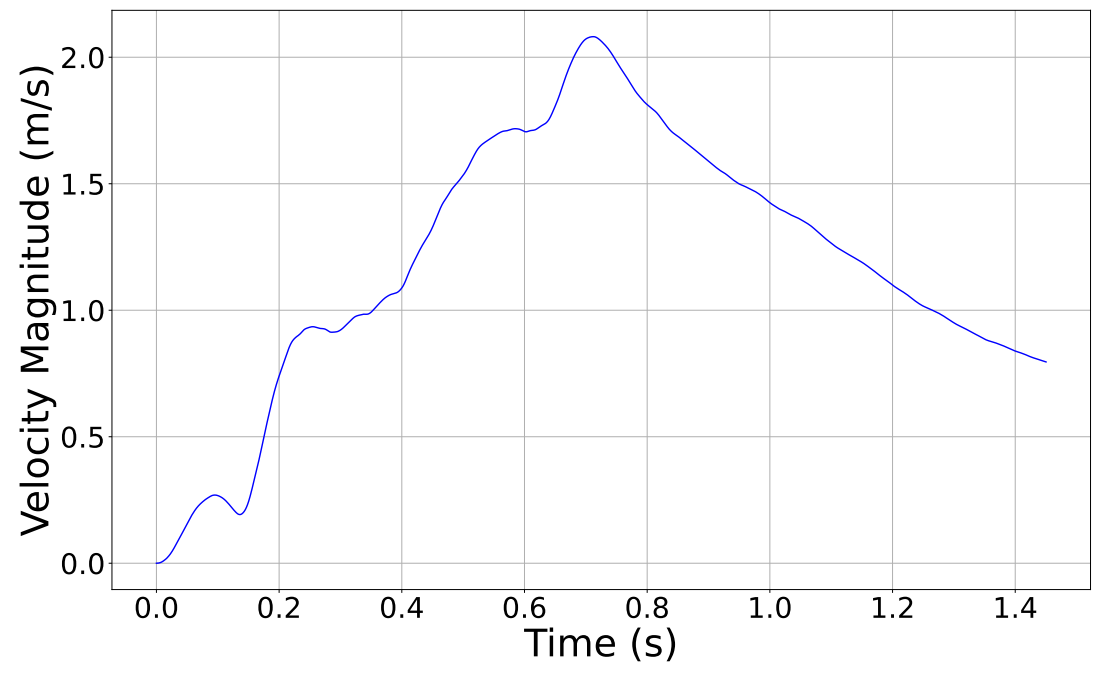

**(b)** Instant velocity magnitude throughout the trajectory

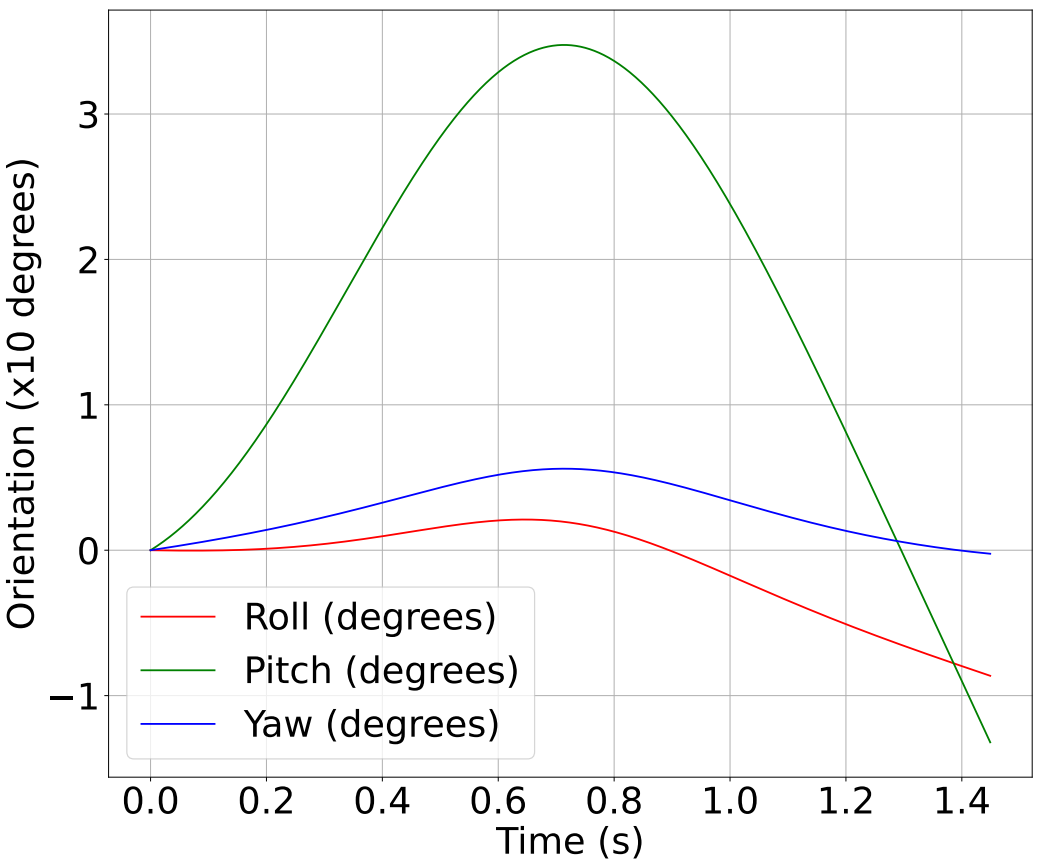

**(c)** Camera orientation throughout the trajectory

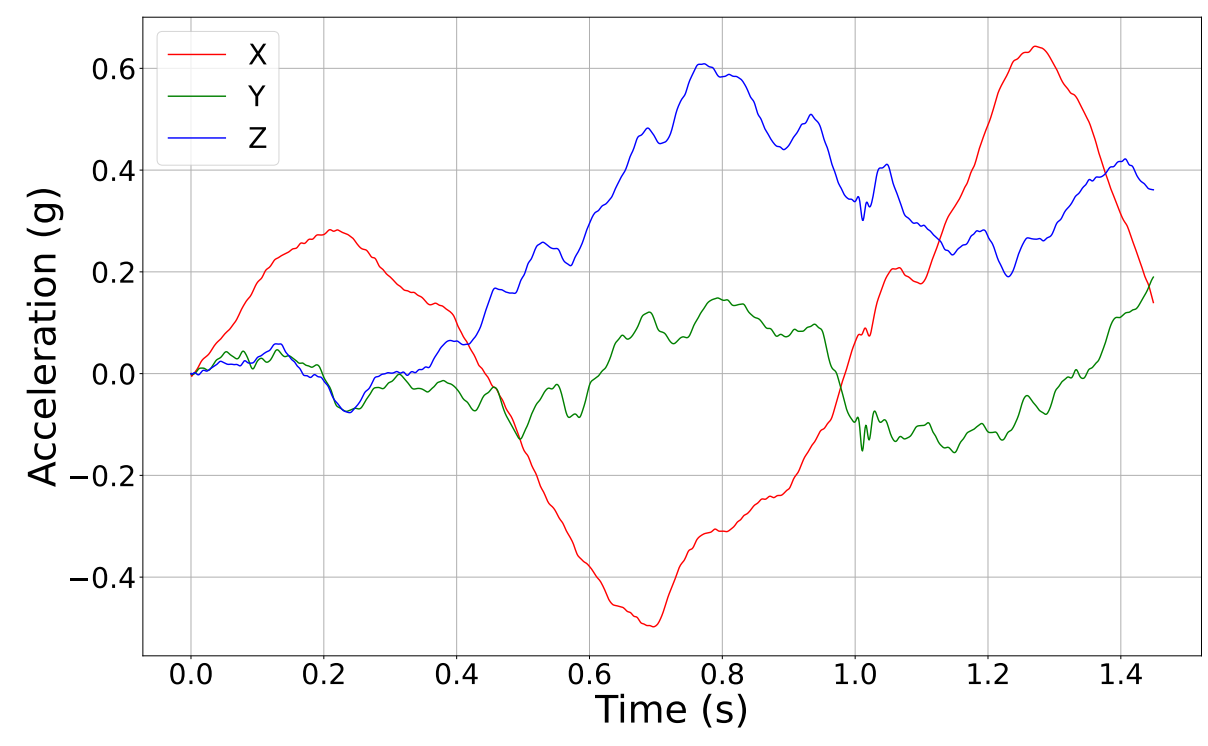

**(d)** Gravity-compensated acceleration throughout the trajectory

**Figure 8.** An illustration of the DAVIS346 camera's 'Z'-shaped trajectory with smooth, flowing bends is shown in (a), depicting the trajectory's two horizontal segments at different distances from the object, connected by a diagonal segment. The trajectory spans a range of 0.2 to 0.7 m, effectively covering this distance. Additionally, the corresponding instantaneous velocity magnitude profile (b), the variation in camera orientation along the trajectory (c), and the gravity-compensated acceleration (when the camera is placed on a horizontal surface pointing towards the object for capture, gravity acts in the negative 'y' direction of the camera's coordinate system) (d) are presented. (In this illustration, the trajectory values are derived solely from the accelerometer and gyroscope data obtained from the DAVIS346 camera's IMU during a single data capture sequence in a laboratory setting and may exhibit minor inaccuracies.)

inside tunnels, the settings specifically used for nighttime conditions were applied.

Following the guidelines in[58], we configured the DAVIS346 biases using the settings detailed in Table 3. For nighttime data collection, we increased the PrBP bias to enhance the camera's sensitivity and reduced the PrSFBP bias to specifically minimize nighttime noise. To prevent excessive motion blur, the PrSFBP bias was carefully adjusted, achieving an balance between noise reduction and image clarity.

After adjusting the biases, we focused on the DAVIS346 using three knobs: 'Near-Far,' 'Open-Close,' and 'Wide-Tele.' For daytime data collection, we set the cameras focus at a 50 cm distance from the object/scene, as the data was collected within a range of 20-70 cm. First, we adjusted the 'Wide-Tele' knob to the 'Wide' setting, then positioned the 'Open-Close' knob in the middle of its range. Finally, we used the 'Near-Far' knob to achieve a sharp event-based visualization in the JAER software. For nighttime data collection, the camera was focused at approximately 40 cm, as the data was collected within the 20-50 cm range. In this case, the focus was adjusted using the 'Near-Far' knob to match the 40 cm distance, following the same procedure as for daytime, but keeping the aperture fully opened.

After setting the biases and adjusting the focus, the camera was calibrated to determine distortion parameters, which were subsequently used to correct the radial distortion of the camera lens[57]. We employed the 'SingleCameraCalibration' filter from the JAER software for both camera calibration and lens distortion correction. The calibrated camera parameters we used are available in the shared code repository.

As already mentioned, we used an energy-efficient IR laser as an external illumination source during nighttime data collection when the ambient lighting was insufficient to capture any meaningful information with either the APS or DVS sensors of the DAVIS346 camera, as observed in the JAER event-histogram visualizer and grayscale frame visualizer. In particular, we used a laser illuminator in night outdoor conditions when the measured ambient lighting is less than $2.5 lux$. At night indoor (inside the laboratory) we

**Table 2.** Data Collection Operational and Environmental Conditions: All numerical values for 'Lighting Conditions' are reported in units of $lux$, measured using the "Light Meter LM-3000 app"[59]. For scenarios involving laser use, the measured laser power using the S120C Standard Photodiode Power Sensor[60] was around 35 mW. In night outdoor scenarios when the measured light level is below $2.5$ lux, the laser illuminator was employed. During night indoor data collection using the laser, measured lighting levels ranged from $0.58$–$0.64$ lux. All the approximate velocity values, expressed in $m/s$, and were derived from the DAVIS346 camera's integrated IMU data. All distance measurements are reported in meters ($m$).

| Parameter | Description | Laboratory (Indoor) Conditions | Field (Outdoor) Conditions |
|---|---|---|---|
| Distance to target | Distance between the camera and the defect surface. | 0.2 - 0.7 (day time)<br>0.2 - 0.5 (night time) | 0.2 - 0.7 (day time)<br>0.2 - 0.5 (night time) |
| Trajectory pattern | The movement paths of the camera in a recording sequence | Z-shaped camera trajectory with smooth, flowing bends. | Z-shaped camera trajectory with smooth, flowing bends |
| Average Speed of movement | Range of average Speed at which the camera is moving relative to the target in a recording sequence | 0.35 - 1.2 | 0.5 - 1.6 |
| Surface texture | Texture of the defect surface | RGB images which shows smooth, regular, irregular, rough surfaces | smooth, regular, irregular, rough surfaces |
| Lighting conditions | Natural or various artificial lighting conditions. | Day , fluorescent(160), laser (0.58 - 0.64) | sunny, cloudy, night time street light (2.5 - 18), laser (<2.5), dimmed fluorescent in tunnel (2 - 12) |
| Environmental Factors | External factors like temperature, moisture, dust, and weather conditions. | 23°C temperature, dry, no-dust | wet, dry, dust, moisture, 1-32°C temperature |

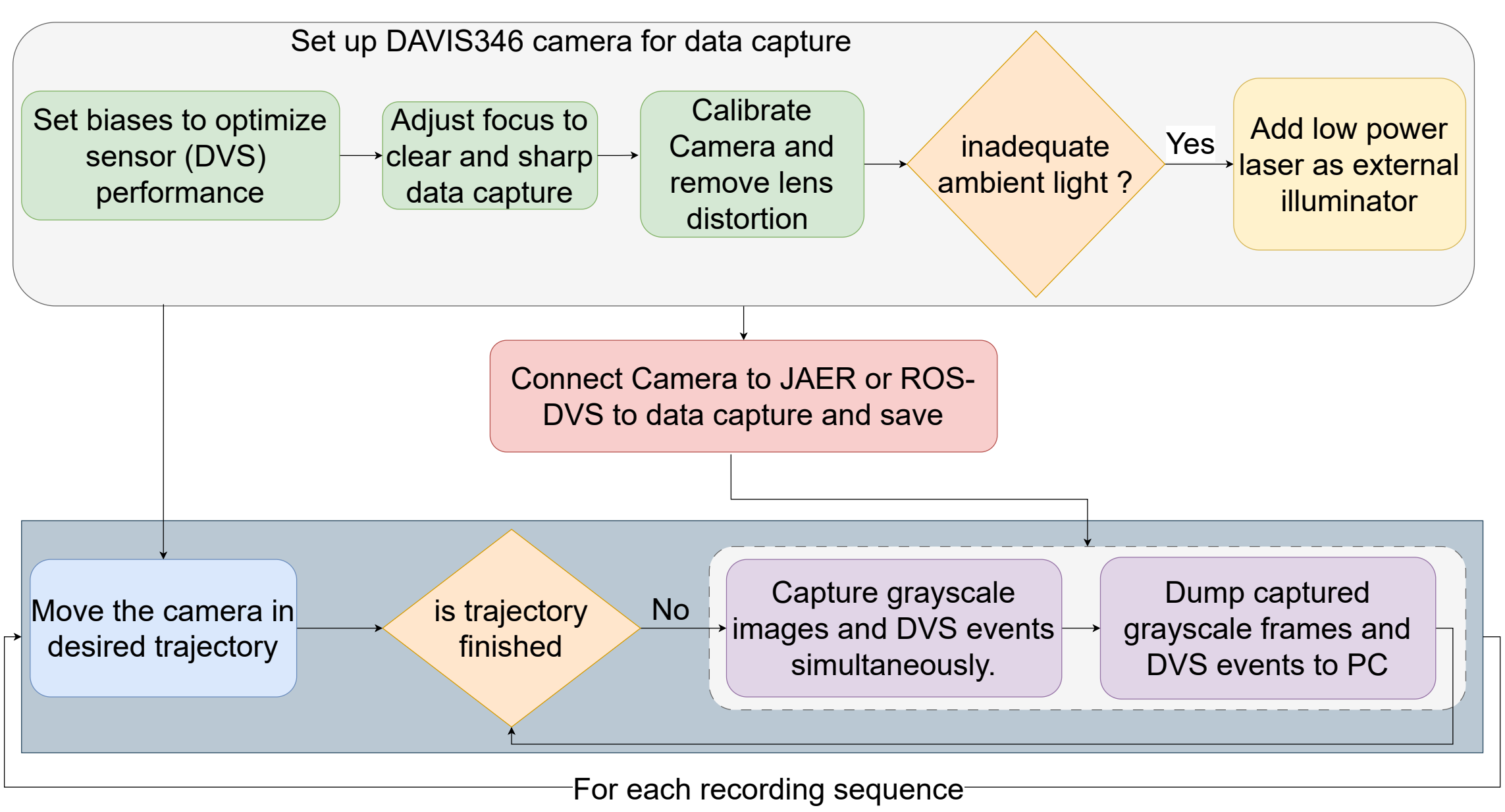

**Figure 9.** Overview of the data collection process consisting of preparing the DAVIS346 camera, capturing grayscale images and DVS events simultaneously, and transferring the data to a PC after each recording sequence.

**Table 3.** Bias settings of DAVIS346 for daytime and nighttime data collection

| | Day | | Night | |
|---|---|---|---|---|
| | c | f | c | f |
| PrBP | 2 | 58 | 3 | 92 |
| PrSFBP | 6 | 16 | 6 | 10 |
| DifBN | 4 | 109 | 4 | 109 |
| OnBN | 5 | 202 | 5 | 139 |
| OffBN | 3 | 53 | 3 | 76 |
| RefrBP | 4 | 25 | 4 | 25 |

switched off all the internal light sources and then employed the laser illuminator. In this situation, the ambient light levels were in the $0.58-0.64 lux$ range due to the different levels of outside lighting coming through the windows on different days. Our motivation for using an IR laser as an external light source stems from the recent emergence of IR laser illuminators as power-efficient alternatives for UAV-based inspections[38]. IR lasers offer additional advantages, such as superior penetration capabilities through dust, smoke, and fog, compared to traditional light sources[39]. These features make them particularly suitable for inspections in

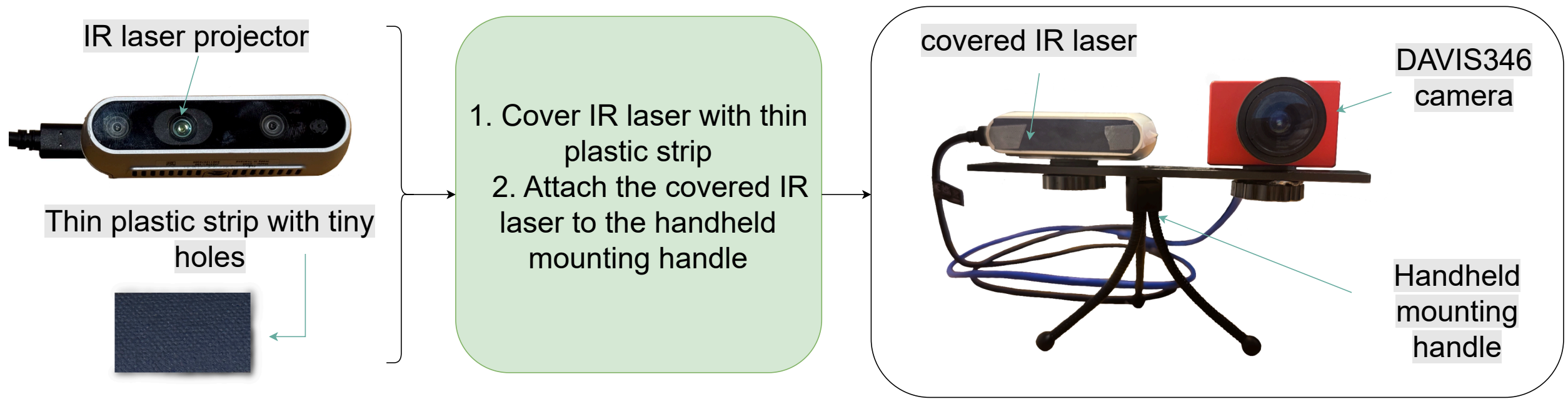

**(a)** Preparation and incorporation of IR laser together with DAVIS346 camera

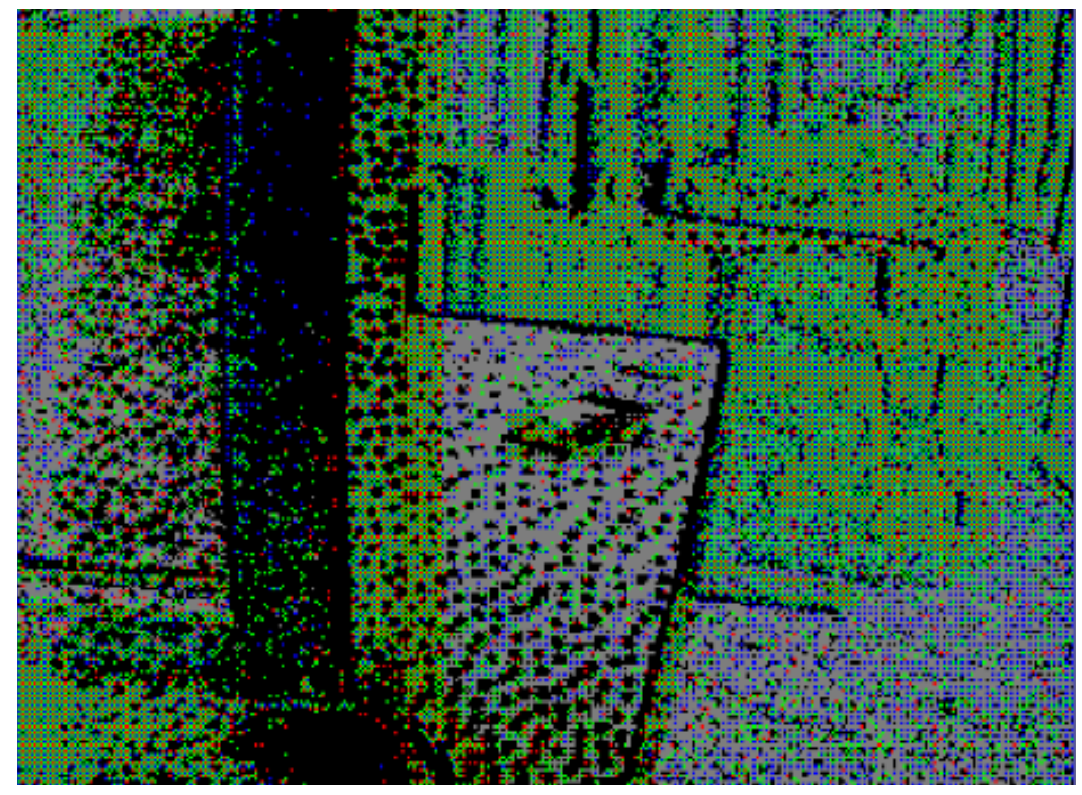

**(b)** Before IR laser is not covered with a thin plastic strip

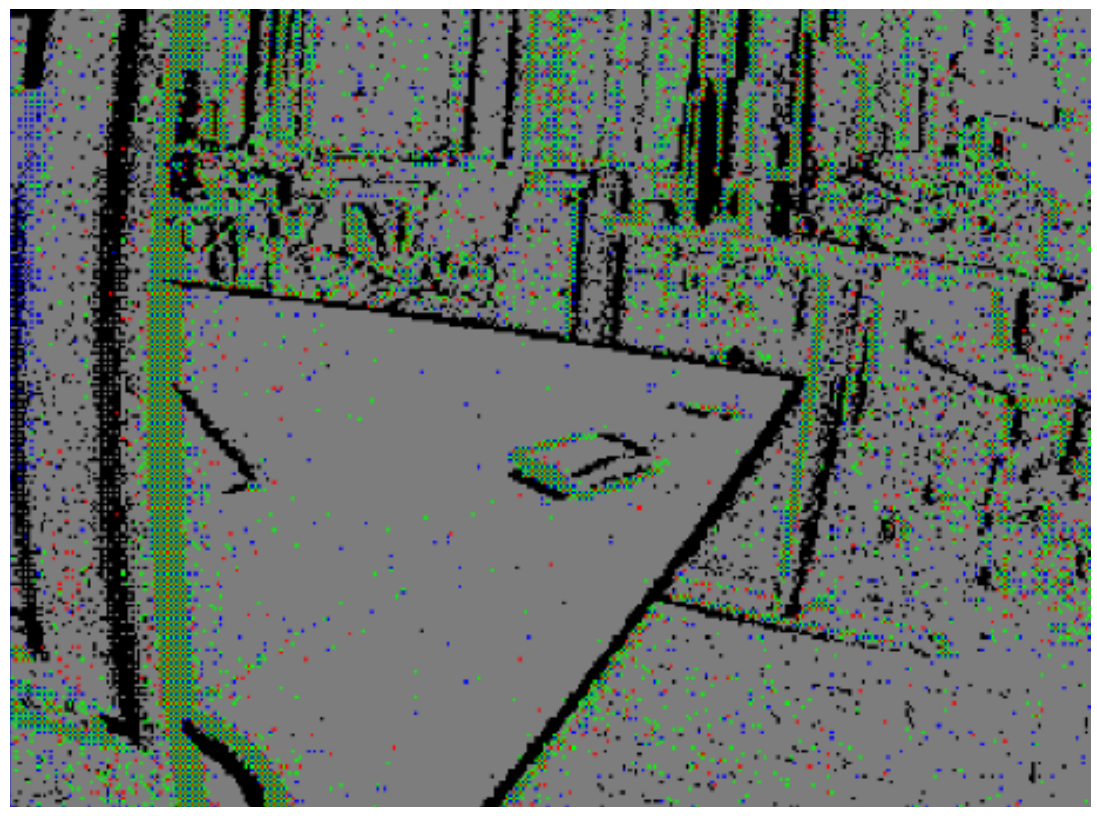

**(c)** After IR laser is covered with a thin plastic strip tunnels or enclosed spaces with limited visibility. In our case, we used the structured laser light available with the Intel realsense D435 camera[37], as shown in Fig. 10a. This IR laser emits beams in a structured block pattern. To prevent the camera from capturing these dot shapes, we covered the laser emitter with a thin plastic strip that had tiny holes, as shown in Fig. 10a. As the diameter of these tiny holes are comparatively smaller than the diameters of projected dots, this strip dispersed the dot patterns while still allowing the laser light to emit effectively. Figures Fig. 10b and Fig. 10c demonstrate the captured scene under the structured laser light, with and without the plastic strip applied, respectively. It can be seen that when the IR emitter was not covered with a thin plastic strip, the structured dot patterns were also captured by the DVS. In this work, the Intel Realsense D435 was only used to employ its IR laser emitter as the external low-power light source. All lighting measurements in lux were taken using the "Light Meter LM-3000 app"[59], available for iPhones. As measured with photodiode power sensor[60], the covered IR laser emitter has 35mW power. Additionally, the measured light levels for other lighting condition scenarios are included in Table 2.

**Figure 10.** The process of preparing and integrating an IR laser as an external illuminator with the DAVIS346 camera for scenarios requiring external illumination. (a) The IR laser projector of the Intel RealSense D435 is covered with a thin plastic strip containing tiny holes, where the hole diameter is smaller than that of the structured dot patterns. The covered laser projector is then attached to the handheld mounting handle along with the DAVIS346 camera. (b) When the IR laser is not covered with a plastic strip, structured dot patterns appear in the captured scene. (c) When the IR laser is covered with a thin plastic strip, the structured dot patterns do not appear in the captured scene.

*Data Capture and Storage* :

We simultaneously captured both event data and grayscale image frames. The captured events exhibited a maximum temporal resolution of approximately 1 ms. During the daytime, grayscale image frames were captured at around 30-35Hz. In nighttime data collection, in scenarios where an external laser light source was used, the APS auto-exposure feature of the DAVIS346 was enabled to automatically adjust the cameras exposure time, which determines how long the APS sensor is exposed to light while capturing a frame. Under low-light conditions, when auto-exposure is enabled, the APS exposure time is increased to gather more light, which in turn reduces the grayscale image frame capture frequency. Consequently, during nighttime data collection, the frame rate varied between 5 Hz and 30 Hz. Both captured events and grayscale image frames were transferred and stored on a PC using either the jAER[44] or ROS-DVS[56] software as shown in Fig. 7. Each recording sequence was organized into separate .h5 files: one for the captured events and another for the captured gray scale image frames. The structure of these .h5 files for events and gray scale images in the given recording sequence is explained in section "ev-CIVIL Dataset Description".

*Data Annotation* :

We annotated our dataset using an approach similar to that in prior event-based object detection studies[26],[27],[28] and[30]. Crack and spalling defects were labeled with rectangular bounding boxes to localize them in 2D event histograms or grayscale images. Each bounding box was assigned a unique integer ID indicating its defect type(whether a crack or a spalling). All bounding boxes adhere to the COCO[46] annotation convention.

For daytime data annotation, the process was based on the grayscale image frames. Since events and images were captured simultaneously, the same label applied to a grayscale image frame was also assigned to the corresponding events. The correspondence between an image frame and events was determined by the proximity of their timestamps. Specifically, if an image frame was captured at timestamp $t$ ms, the events in the temporal window $[t-10, t+10]$ms were considered as corresponding to that particular image frame. Thus, the annotations for the image frame were also applied to those events. During daytime, the annotation frequency was approximately 30 Hz, as annotations were based on grayscale image frames, which were captured at about 30 Hz.

In the annotation process for grayscale image frames within a given recording sequence, we initially annotated a set of randomly selected frames manually using the "LabelMe" software[45]. Then, we used the OpenCV object tracker[61] to propagate these annotations to unlabelled frames. Finally, all the frames were reviewed, and any necessary corrections were manually applied.

In the case of nighttime data labeling, when the grayscale frames were underexposed or insufficiently illuminated, making them non-informative to the human eye, we relied on the captured event data from the DVS sensor for the annotation process. In such cases, we selected events within a 40 ms temporal window and generated a 2D event histogram using the algorithm 1. The annotation was then based on this 2D event histogram. Figure 2.b shows an example of a crack annotation performed using such a 2D event-histogram frame. The event-histogram-based annotation was performed at a frequency of 15 Hz. Additionally, when the grayscale image frame rate dropped below 15 Hz due to the APS auto-exposure feature, which increased the APS visibility at night, intermittent 2D event histograms were generated and annotated. When the grayscale image frames are adequately illuminated, allowing information to be visible to the human eye, we used the same grayscale-based annotation convention as in the daytime data annotation.

Generally, if the annotation is based on a grayscale image frame, the timestamp corresponds to the timestamp of the event which is closest to the timestamp of that frame. However, if the annotation is based on a 2D event histogram, the timestamp refers to the time of the event that falls in the middle of the temporal window used to create the histogram, which encompasses all the events that contributed to it.

## Formation of 2D two-channel histograms from DVS event Data

We constructed 2D, two-channel event histograms. The histogram is organized such that the bins form a (346, 260) grid, which corresponds to the spatial resolution (346x260) of the DAVIS346 camera. For each bin in the (346, 260) grid, the number of positive and negative events occurring within a specified temporal period are separately summed. This results in two channels per bin: one channel represents positive events, and the other represents negative events. Positive events refer to those triggered by pixel intensity increments, while negative events correspond to those triggered by perceived pixel intensity decrements. The Eq. (1) shows how the histogram value for the bin $(x_i, y_i)$ is calculated using events that occurred during the period $[t - T/2]$ and $[t + T/2]$ for the channel corresponding to positive events.

Fig. 5 visualizes a 2D event histogram formed using the event volume corresponding to events collected over a $15ms$ temporal period by the DVS sensor of the DAVIS346 camera. Based on this visual representation, it is evident that the collected events have the potential to successfully capture the scene, which includes a spalling defect.

$$h_{pos}(x_i, y_i) = \sum_{t-T/2}^{t+T/2} I_{pos}(x_i, y_i) \tag{1}$$

The temporal period $T$ in the equation can be determined either by a fixed time interval, such as 15ms, 30ms, and so on[62], or by a specific number of events, such as 5000 events, 10000 events, and so on[62].

As shown in Fig. 11, the visual quality of the 2D event histograms depends strongly on the time window used to generate them. Spalling typically requires a longer time window, but applying the same window to cracks can cause them to appear blurred. Even different types of cracks may require different time windows to improve the visual clarity of the scene without causing blurring or disappearance. The same principle applies to spalling as well. To address this, we adopt an adaptive temporal length strategy for 2D histogram formation. Initially, we considered event density based methods[?] , in which a predefined number of events are accumulated before constructing the histogram, allowing the temporal length to vary. However, in practice, noisy spikes from the DVS sensor can cause the event count to reach the threshold prematurelybefore the scene content is adequately represented.

To mitigate this issue, prior work[63] proposed dividing the pixel array into a spatial grid (e.g., 44) and collecting events until any grid cell receives a predefined number of events, thereby dynamically adjusting the temporal window. Nonetheless, we observed that widespread sensor noise can cause multiple grid cells to meet this condition too early, again resulting in poorly formed scene representations.

To overcome these limitations, we introduce a new adaptive event-selection algorithm (see Algorithm 1). In this approach, the deviation of each grid cells event count from the mean event count across all grid cells is monitored. When this deviation exceeds a specified threshold and the temporal length surpasses a minimum empirically determined threshold, the corresponding events are used to form the 2D event histogram. As a result, the algorithm automatically adjusts the temporal window based on scene dynamics, producing variable-length event segments that more accurately capture the defects which we experimentally validated in the results section in terms of detection performance improvement.

In Algorithm 1 we select an event volume around any event with time stamp $t$. If this event has id $I_{id}$ in the source event-based data stream, then event volume is selected around this this event, so that temporal length of the event selected event volume exceeds a predefined time window threshold $T_{th}$ and the event count in at least one grid cell of an $(n, m)$ grid over the pixel space surpasses a threshold $A_{th}$ relative to the average event count of all grid cells.

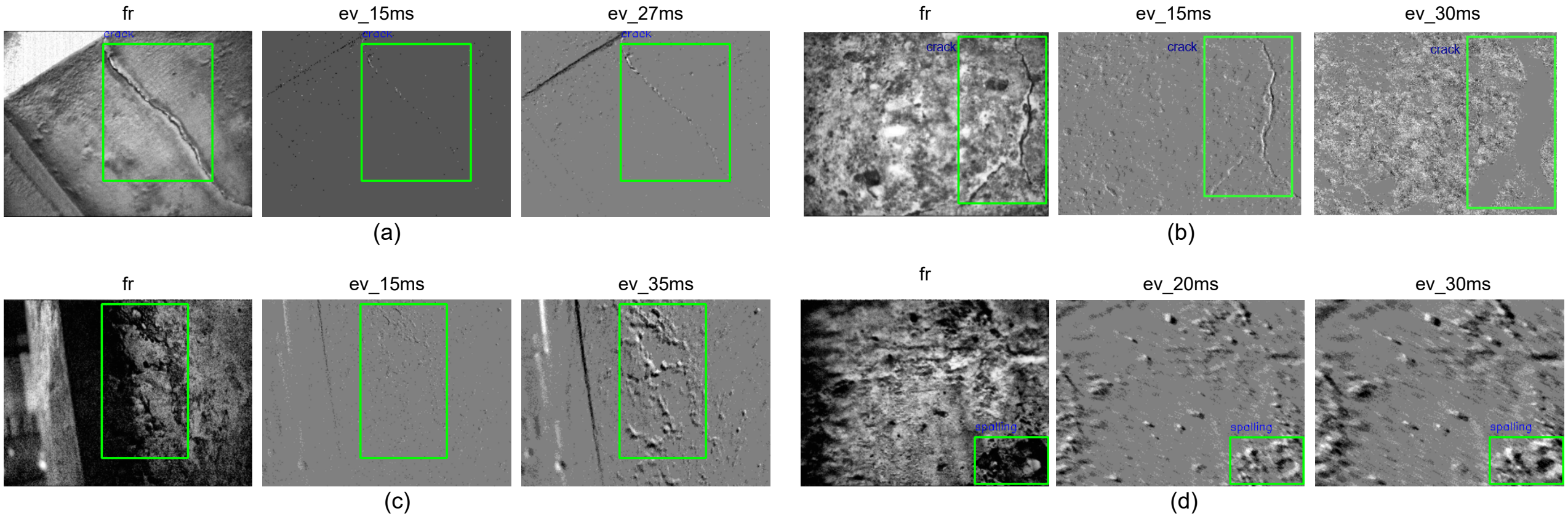

**Figure 11.** Influence of Temporal Window on 2D Two-Channel Event Histogram Generation where (a) A crack visible with T = 27ms but unclear with T = 15ms (b) A crack visible with T = 15ms but unclear with T = 30ms (c) A spalling clearly visible with T = 35ms but not clearly visible with T = 15ms (d) A spalling clearly visible with T = 20ms but blurred with T = 30ms (fr : grayscale image frame $ev_T$ 2D event histogram with temporal length 'T')

For our case we found that $(4, 4)$ and 175 as the $T_{th}$. The events were incrementally collected to the desired event volume by adding $q \times count$ events around it. We used 100 as the $q$ for our case. $count$ is the counter which is incremented by 1 each time a new $2q$ events are added. After each addition of $2q$ events, the algorithm check whether the selected event volume exceed the $T_{th}$ and $A_{th}$ and if it is, those selected event volume is used to create the 2-channel 2D event histogram as explained in Algorithm 1. Based on our empirical study, We used $15ms$ as the $T_{th}$ for daytime or adequately illuminated data. However, for data collected under low-light conditions such as night outdoor settings, tunnel environments, and night indoor scenarios with laser lighting a higher temporal length was required compared to well-lit data, because in these data, the number of non-noisy events per unit of time is less compared to well-lit data. So for those data we used $30ms$ as the $T_{th}$.

Fig. 12 shows the distribution of temporal lengths of the event volumes used to form 2D histograms in the test set. It can be observed that the temporal lengths range from 15 ms to 100 ms.

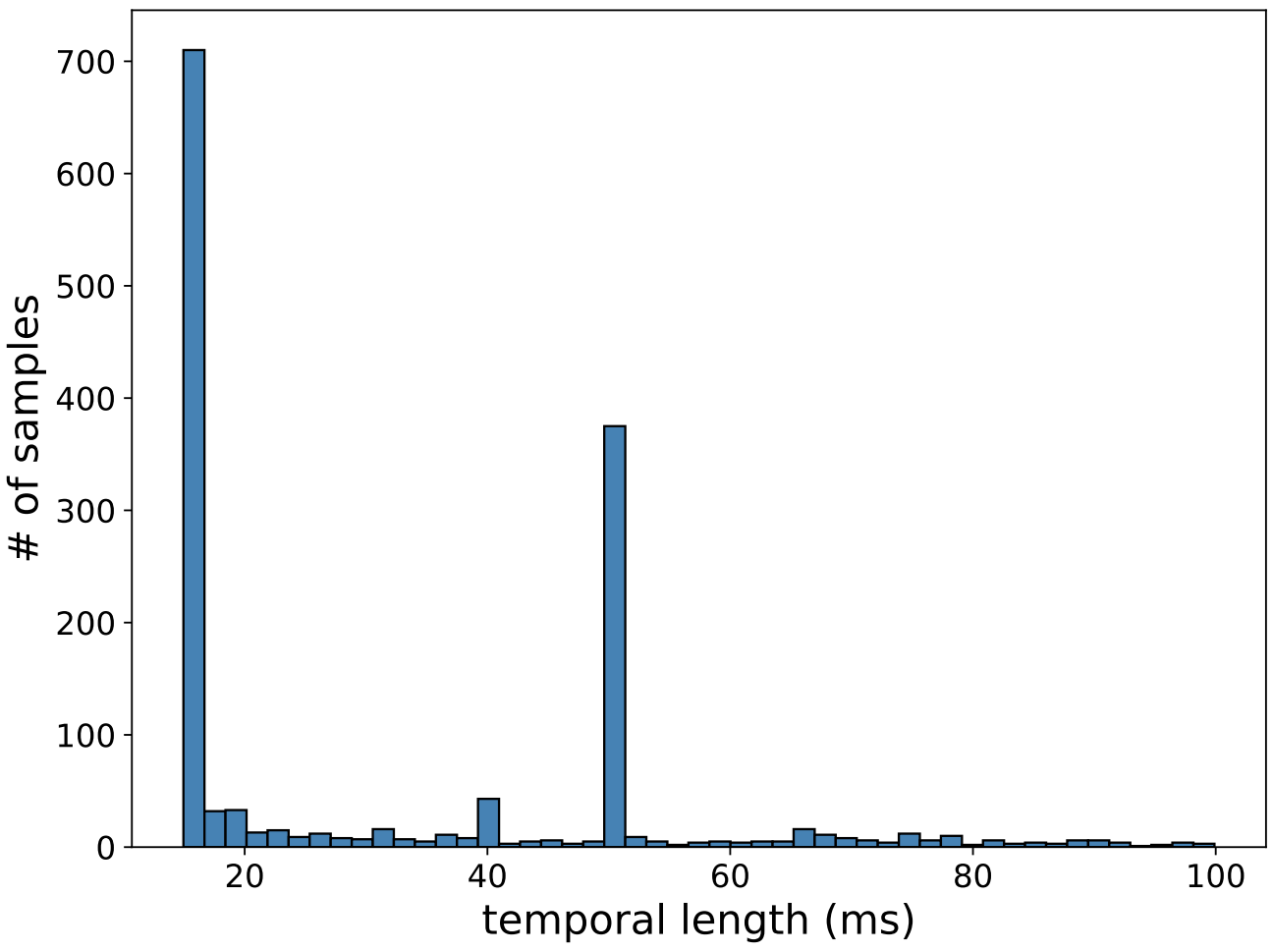

**Figure 12.** Distribution of temporal lengths (time windows, in ms) of the test set event volumes used for 2D 2-channel histogram formation, obtained using the method described in Algorithm 1.

## *Deep learning architectures for defect detection*

We benchmarked our dataset using four real-time object detection models: YOLOv6m[16], YOLOv6lite-s[16], SSD300 with a ResNet-50 backbone[16,19], and SSD300 with a MobileNetV2 backbone[16,49]. Among these, YOLOv6lite-s and SSD300-MobileNetV2 are lightweight architectures specifically designed for deployment on mobile or edge devices.

Both YOLOv6-based models consist of a backbone network for feature extraction, a neck component that fuses multi-scale features from different stages of the backbone, and an anchor-free detection head that localizes and classifies crack and spalling defects at multiple spatial resolutions. In contrast, the SSD models use anchor-based detection. SSD300 with a ResNet-50 backbone relies on a deeper and more parameter-heavy feature extractor, while the SSD300-MobileNetV2 variant uses MobileNetV2, which has substantially fewer layers and parameters, resulting in significantly lower computational cost. The SSD detection head operates directly on feature maps extracted at multiple backbone levels, with each level dedicated to detecting defects of different scales, and without any additional feature-fusion mechanism.

More detailed architectural descriptions of these models are provided in the Appendix.

Our choice of YOLOv6, an anchor-free object detector, is consistent with prevIoUs literature, where YOLOv6 and YOLOv8 have demonstrated strong performance on event-based object detection[30,73]. Anchor-based detectors such as SSD have also been explored for event-based detection in prior studies[72]. Both YOLOv6 and SSD are capable of real-time inference, which is a major requirement for our application.

Transformer-based architectures (e.g., ViT, DETR), despite their success in frame-based object detection[79,80], were not considered for our study due to their high computational cost, which makes real-time performance challenging. Moreover, these models require large amounts of event-based labeled training data. Recent studies[78] indicate that transformers have not yet outperformed CNN-based architectures for event-based detection. This is largely

**Algorithm 1** Algorithm to select an event volume having adaptive temporal length and create a 2-channel 2D event histogram from the selected event volume: The event volume is selected around a user-preferred anchor event with an identifier $I_{id}$. The selection ensures that the temporal length of the event volume exceeds a predefined time window threshold $T_{th}$ and the event count in at least one grid cell of an $(n, m)$ grid over the pixel space surpasses a threshold $A_{th}$ relative to the average event count of all grid cells.

**function** SELECT_EVENTS_AND_CREATE_HIST(*ev_org*, $T_{th}$,$A_{th}$,*I_id*)
    *ev_org* ▷ original events volume as an array(t,x,y,p)
    $T_{th}$ ▷ time threshold to select event volume
    $A_{th}$ ▷ area threshold to select event volume
    *q* ▷ packet size to incrementally form desired event volume
    *count = 1* ▷ counter the number of packets *q*
    *m,n* ▷ grid dimension for area event count calculation
    $I_{id}$ ▷ event index around which the desired event volume should be formed
    **while** True **do**
        $s_{id} = I_{id} - q \times count$ ▷ desired start event id
        $e_{id} = I_{id} + q \times count$ ▷ desired end event id
        *ev_desired= ev_org[s_id:e_id]* ▷ desired events

        Form 2D event histogram for (m,n) grid *h_m_n_arr*
        Calculate the mean of the (h_n_m_hist2d) *mean*

        $extra_events_per_cell = h_m_n_arr - mean$ ▷ Calculate this difference for each grid cell in (m,n) grid

        $current_time_window = ev_org[e_{id}, t] - ev_org[s_{id}, t]$ ▷ Calculate the temporal length of the selected event volume
        $count = count + 1$

        **if** $(extra_events_per_cell > A_{th}$ and $current_time_window > T_{th})$ **then**
            $candidate_ev_vol_found = True$
            ***break***
        **end if**
    **end while**

    **if** $candidate_ev_vol_found$ **then**
        Create 2D event histogram *h_x_y_ev* with selected event volume *ev_desired* using Eq. (1)
        Clip the *h_x_y_ev* to the range of $3 \times std$ ▷ *std* standard deviation over all the bins in histogram
        Normalize the clipped *h_x_y_ev* by diving its values by its maximum value
        **return** clipped and normalized histogram *h_x_y_ev*
    **end if**
**return**
**end function**

due to the inherent incompatibility with event data where the sparse and asynchronous nature of spikes means that many input patches contain little to no information.

*Classification Architectures* The standalone classification performance was evaluated using four classification architectures: ResNet34[19], VGG16[18], EfficientNet-b0[52], and MobileNetV2[49]. In all these models, a sequence of convolutional and pooling layers is employed to extract features and gradually reduce the spatial resolution through the feature maps across layers. These feature extraction and spatial resolution reduction layers also serve as the backbone of object detection network architectures.

In addition to these feature extraction layers, the classification architectures incorporate fully connected layers for feature fusion. The fused features are then passed to the output layer of the classification architectures, which determines a discrete probability distribution over the set of classes to be predicted (e.g., crack or spalling class). A softmax operation is applied to compute this probability distribution, and the class with the highest probability is selected as the predicted class to which the input data is classified.

### *Evaluation metrics*

To assess defect detection performance, we used well-known and widely used mean Average Precision (mAP) metrics which are initially defined alongside the COCO dataset in[53]. More specifically, we use the following three metrics.

- $mAP_{0.5}$ metric which is the mAP at an IoU (Intersection over Union) threshold of 0.5, representing the average accuracy of object detection across all classes based on how well predicted bounding boxes overlap with ground truth bounding boxes.
- $mAP_{0.5:0.95}$ is the mean Average Precision computed across a range of IoU thresholds from 0.5 to 0.95, indicating the average accuracy of object detection considering various levels of overlap between predicted and ground truth bounding boxes.
- $F1_{\text{IoU}0.5}$ metric, which combines precision and recall to provide a balanced assessment of the models ability to both correctly classify and accurately localize defects. This metric is computed at an IoU threshold of 0.5, meaning a prediction is counted as correct only when it has the right class label and overlaps sufficiently with the ground-truth bounding box.

To implement these detection metrics, we leveraged the APIs provided by the pycocotools library[54].

An in detailed description on these metrics are included in the appendix.

While the $F1_{IoU0.5}$ metric offers an evaluation of classification performance, its dependency on detection performance can be noted. Therefore, for a standalone assessment of classification performance, we employ the widely utilized accuracy metric, defined in equation Eq. (2). In Eq. (2), $P_{correct}$ represents the number of correctly classified samples, while $P_{total}$ denotes the total number of samples utilized for testing. The accuracy metric quantifies the percentage of correctly predicted class labels among all samples in a dataset.

$$accuracy = \frac{P_{correct}}{P_{total}} \tag{2}$$

## Setup for Benchmarking

In this section, we describe the process of extracting data samples from the recorded field and laboratory sequences, the division of the data into training, validation, and test sets, as well as the training methodology and testing approach for the deep learning architectures.

### *Extraction of data samples from sequences*

A detailed illustration of the selection of data samples for the dataset benchmarking process is shown in Fig. 13. To benchmark the dataset, data was extracted from all recording sequences. For each sequence, approximately 1015 samples were randomly selected.

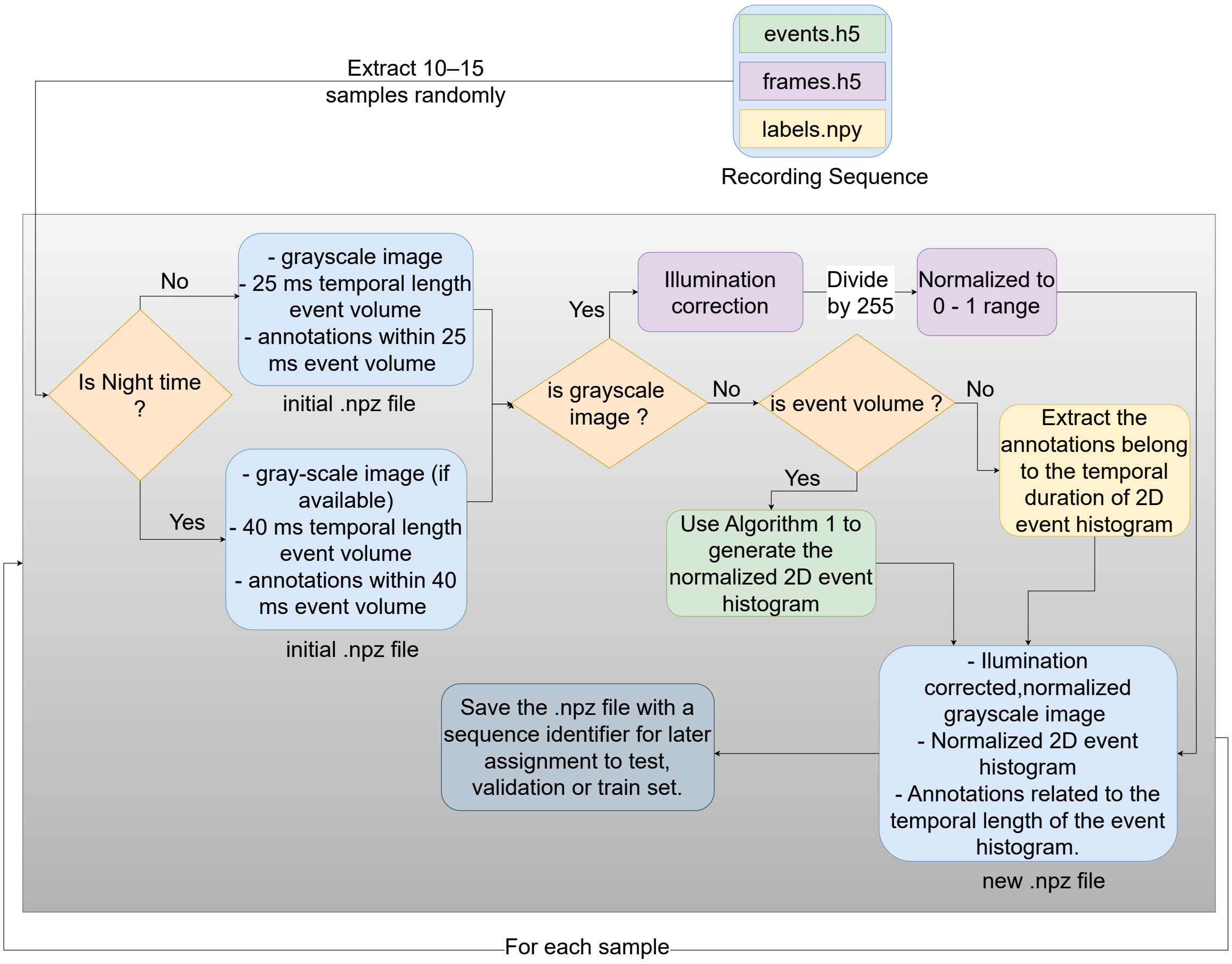

**Figure 13.** Extraction of grayscale images and event-based data from the evCIVIL dataset for benchmarking crack and spalling detection. First, 10-15 samples are obtained from each recording sequence. For each sample, 2D event histograms are generated from the corresponding events. The extracted grayscale image frames are then preprocessed. These preprocessed grayscale images, 2D event histograms, and extracted annotations are saved together in .npz files, which will be used to create training, testing, and validation sets.

Each daytime data sample typically includes the following components:

- A grayscale image frame.
- A 25ms temporal length event volume.
- Bounding box annotations corresponding to the grayscale image, which are same for the event volume.

Nighttime samples typically share similar components with daytime samples. However, when the APS auto-exposure is enabled, there can be instances where a direct one-to-one correspondence between a 2D event histogram and its associated grayscale image frame does not exist. This discrepancy arises because the capture frequency of the grayscale images drops below 15 Hz when the APS sensor's exposure time is extended in low-light conditions. As a result, some samples in the sequence only include the 2D event histogram and the corresponding bounding box annotations.

All grayscale images selected were corrected for illumination using OpenCV's Contrast Limited Adaptive Histogram Equalization (CLAHE) filter[48] and their intensity values were normalized to a range of 0 to 1. From the selected event volumes, 2D event histograms were generated using Algorithm 1. Subsequently, annotations corresponding to the temporal range of these events were extracted. These extracted annotations generated 2D event histograms, and grayscale images (where available) were then stored in a .npz file format to enable efficient storage and retrieval. This approach facilitated the selection of 2,280 samples from field data and 4,320 samples from laboratory-collected data for the benchmarking process.

## *Train, Validation and Test Sets*

The main advantage of DVS cameras, which trigger events, over APS cameras, which provide grayscale frames, is their ability to operate effectively in low-light and dynamic lighting conditions. To validate this capability of DVS cameras in the context of visual inspection of civil infrastructure defects, we created two test sets using 183 field data sequences. One test set, called the "Low-Light Test Set", includes samples collected under low, dimmed, or dynamic lighting conditions, while the other, named as the "Adequately Illuminated Test Set", comprises samples gathered under adequate or well-lit conditions.

The Low-Light Test Set consists of samples extracted from 70 sequences, which include defects captured in outdoor nighttime conditions (laser and street lighting), saturated/over lighting, and indoor dim lighting such as inside the tunnel. These sequences contain 89 unique occurrences of cracks and 26 instances of spalling. Adequately Illuminated Test Set consists of samples extracted from 113 sequences of well-illuminated daylight

**Table 4.** Number of samples used for train, validation and test sets

| Train | Validation | Test | |
|---|---|---|---|
| | | Adequately Illuminated | Low-light |
| 4900 | 550 | 1556 | 725 |

conditions such as sunny, cloudy, dark, and indoor fluorescent light, which consists of 68 distinct instances of spalling and 163 unique instances of cracks. In this case, sample extraction for each sequence was done following the approach explained in relation to the

Fig. 13.

The remaining 115 sequences from the field, which contain 206 distinct cracks and 27 distinct spalling instances, were used together with 360 laboratory sequences, which contain 220 distinct cracks and 308 spalling instances, to form the train and validation sets. In this case, as well, samples from each sequence were extracted using the approach explained in relation to the Fig. 13. Since 10 to 15 samples were extracted from each sequence, all together 550 samples, representing data collected under various lighting conditions, were allocated to the validation set. The remaining 4900 samples were used for the training set. The number of samples used for train,validation and test sets are included in Table 4.

### *Training Deep Learning Architectures*

All deep learning architectures were implemented and trained using the PyTorch framework[50]. Before starting the training, all the detection architectures were initialized with COCO pre-trained weights. Grayscale images or 2-channel event 2D histograms are spatially resized to 640 $\times$ 640 from the initial 346$\times$260 spatial dimension before feeding to YOLOv6m. During this process, we ensure that the aspect ratio of the original images is preserved to avoid distortion of the object shapes. Following the same resizing conventions, the grayscale frames and 2-channel event histograms were spatially resized to 300 $\times$ 300, before being fed into the SSD detection architecture. The confidence threshold during training was set to 0.02, while during testing, it was increased to 0.2. In both training and testing, IoU threshold for Non-Maximum Suppression (NMS) was set to 0.4.

To train the classification architectures, the input data consisted of patches within the bounding boxes localizing the crack and spalling defects. These patches had varying spatial dimensions. Therefore, for each architecture, we performed separate training by resizing the patches to different input resolutions, such as 32$\times$32, 64$\times$64, 128$\times$128, and 224$\times$224. The classification accuracy was evaluated to identify the best input resolution that works across all patches as a whole. All detection architectures were trained for 400 iterations with an initial learning rate of 0.001, using the SGD optimizer combined with a CosineAnnealingLR scheduler. The same learning rate, optimizer, and scheduler were used to train the classification architectures for 100 epochs.

We applied affine transformations, along with horizontal and vertical flips, as data augmentation techniques.

During training, the three best models from each architecture were saved based on the highest validation $mAP_{0.5}$ values. For testing, these top models from each architecture were first evaluated on the adequately illuminated test set. The model with the highest $mAP_{0.5}$ on the adequately illuminated test set was then used to evaluate the Low-Light test set. For classification architectures, we used the validation accuracy instead of the validation $mAP_{0.5}$ to select the best models for each architecture.

## Experimental Results

During the benchmarking of our ev-CIVIL dataset, we conducted the following experiments.

- The first two experiments evaluate the event-based detection performance of the selected deep learning models in comparison to the grayscale image-based detection performance. These evaluations were conducted separately for the Adequately Illuminated Test Set and the Low-Light Test Set
- In the third experiment, we examined the use of laboratory data to address the scarcity of field data in training deep learning algorithms. This experiment aimed to assess how effectively the learned features of defects from laboratory data generalize to detecting defects in field data.
- The fourth experiment focused on evaluating the significance of our event-based histogram formation method, explained in Algorithm 1, in comparison to the fixed temporal length-based event histogram formation method, in terms of defect detection performance.
- For the last experiment, we cropped the defect areas localized by the bounding boxes of ground truth annotations and used them to evaluate the crack and spalling classification performance using the selected classification architectures.

Table 5 and Table 6 present the quantitative results to assess the detection performance where 'All' refers to the overall metric values when considering both 'Crack' and 'Spalling' defect classes together. Additionally, 'Crack' indicates the detection performance of the 'Crack' defect class, while 'Spall' denotes the detection performance of the 'Spalling' defect class.

For the qualitative visualization of detection results presented in Fig. 14 and Fig. 15, frame-based detections are overlaid on grayscale intensity frames, while event-based detection results are illustrated on 2D event histograms generated using algorithm 1.

### *Detection performance on Adequately-Illuminated test dataset*

The performance of the selected object detection models on the adequately illuminated dataset is summarized in Table 5. According to the table, YOLOv6m achieves the best performance across all three reported metrics for both frame-based and event-based data. The results also show that frame-based detection outperforms event-based detection by a margin of up to 0.10 across all metrics for data collected under proper illumination levels. Additionally, when comparing class-wise detection performance, crack detection consistently surpasses spalling detection in almost all cases. This discrepancy may be attributed to the significantly larger number of crack samples from the field available in the training set compared to spalling samples. Moreover, both YOLO models outperform the SSD models in detection performance for both frame-based and event-based data. This indicates that YOLOv6 models are highly optimized specifically for frame-based image data while still achieving higher event-based detection performance compared to SSD models.

**Table 5.** Event-based and Image-based defect detection performance on Adequately-Illuminated test dataset

| | $mAP_{0.5:0.95}$ | | | | | | $mAP_{0.5}$ | | | | | | $F1_{IoU0.5}$ | | | | | |
|---|---|---|---|---|---|---|---|---|---|---|---|---|---|---|---|---|---|---|
| | All | | Crack | | Spall | | All | | Crack | | Spall | | All | | Crack | | Spall | |
| | fr | ev | fr | ev | fr | ev | fr | ev | fr | ev | fr | ev | fr | ev | fr | ev | fr | ev |
| YOLOv6m | **.31** | .25 | **.39** | .28 | **.23** | .20 | **.53** | .45 | **.59** | .48 | **.46** | .43 | **.60** | .55 | **.65** | .57 | **.55** | .53 |
| YOLOv6lite-s | .27 | .19 | .31 | .20 | .23 | .19 | .50 | .42 | .56 | .44 | .44 | .42 | .56 | .51 | .62 | .50 | .53 | .52 |
| SSD300-resnet50 | .19 | .18 | .24 | .18 | .15 | .17 | .43 | .42 | .47 | .40 | .45 | .39 | .55 | .50 | .57 | .50 | .53 | .50 |
| SSD300-mobilev2 | .14 | .13 | .17 | .12 | .12 | .14 | .32 | .32 | .37 | .35 | .29 | .28 | .46 | .43 | .51 | .45 | .49 | .43 |

Based on the qualitative visualizations in Fig. 14, YOLOv6m demonstrates superior detection quality compared to the other models. In these visualizations, we compare 12 frame-based (fr) and event-based (ev) sample pairs for YOLOv6m and YOLOv6s-lite to illustrate the detection of distinct defect shapes on different surface textures of civil infrastructure. For the first 6 pairs, we also include SSD models for visual comparison of detection quality across the models.

In the first fr-ev pair of the third row of Fig. 14, a spalling and two cracks are detected simultaneously, while the second fr-ev pair of the same row shows successful detection of a small concrete removal as spalling. Also, the first fr-ev pair of the fourth row highlights a scenario where event-based detection captures a full crack, whereas frame-based detection detects only a part of it. In the third fr-ev pair of the fourth row, a spalling is detected on a surface fully covered with cracks. In this case, both the spalling and surrounding cracks are detected as a spalling defect by the model.

Lightweight models show reduced detection quality compared to standard YOLOv6m or SSD300 with ResNet50 backbone models. Occasionally, spallings are misclassified as cracks, as shown in the event-based histogram of the first pair in the seventh row of Fig. 14. However, in most cases, the models successfully localize cracks while ignoring distracting features such as floor tiles or pavement edges.

### *Detection Performance on Low-Light Test Set*

When assessing the detection performance of the models on the Low-Light Test Set, a consistent trend emerges where event-based detection surpasses frame-based detection across all four models, as illustrated in Table 6. Upon examining the $mAP_{0.5}$ and $F1_{IoU0.5}$ metrics, it becomes evident that both the overall and class-wise event-based detection performances exhibit a .20 to .30 improvement over frame-based detection performance. Additionally, the $mAP_{0.5:0.95}$ metric demonstrates a 0.08 to 0.15 enhancement in comparison to its frame-based counterpart.

Fig. 15 shows a qualitative visualization of the detection results on the Low-Light Test Set with some selected cracks and spallings for YOLOv6m and YOLOv6lite-s models. It shows the detection results for both frame-based and event-based detections highlighting the cases where frame-based detection fails and event-based detection succeeds. In Fig. 15, the first two rows depict results for night-outdoor data samples collected using a low-power, low-intensity laser as the source of illumination. The third and fourth rows display the detection results of data samples collected inside a low-illuminated tunnel. The last two rows illustrate results for data captured under other scenarios such as dynamic light, and fast movement under low-light and saturated light conditions. These qualitative results, showcasing the effectiveness of event-based detections compared to the failures observed in frame-based detections, further emphasize the importance of event-based detection in challenging lighting conditions.

### *Leveraging Laboratory Data to Compensate for Limited Field Data: Lab-to-Field Generalization*

Given the limited amount of data collected under field conditions and the fact that most of the available field data has been used for our test sets, we lack sufficient samples for training. Therefore, we utilized a laboratory dataset along with a small number of field data samples that do not belong to our test sets as our training set.

So, in this experiment, we demonstrate the generalizability of detection algorithms trained primarily on laboratory data, supplemented by a handful of field samples, as evidenced by increased detection accuracy. Figure 6 illustrates the differences in detection performance metrics ($mAP_{0.5}$, $F1_{IoU0,5}$) when training with only field data not included in the test set versus when training with both this field data and laboratory data.

Fig. 16 shows the percentage reduction in $mAP_{0.5}$ and $F1_{IoU0.5}$ metrics relative to the corresponding values reported in Table 5 and Table 6, when the selected four detection models were trained without the laboratory data component. Across all four models, an overall performance drop in the range of $15\% - 35\%$ is observed. Among the two defect types, spalling is the primary contributor to the performance drop, with a percentage reduction ranging from $63\% - 95\%$. This significant decline is due to the insufficient number of spalling samples from the field in the training set, which hinders the models ability to effectively learn the features associated with spalling defects. On the other hand, this also demonstrates that the features learned from the laboratory data related to spalling defects are successfully generalized to identify spalling defects in field data. Additionally, the observed drop in crack detection performance emphasizes that laboratory data related to cracks increases the useful amount of crack-related data in

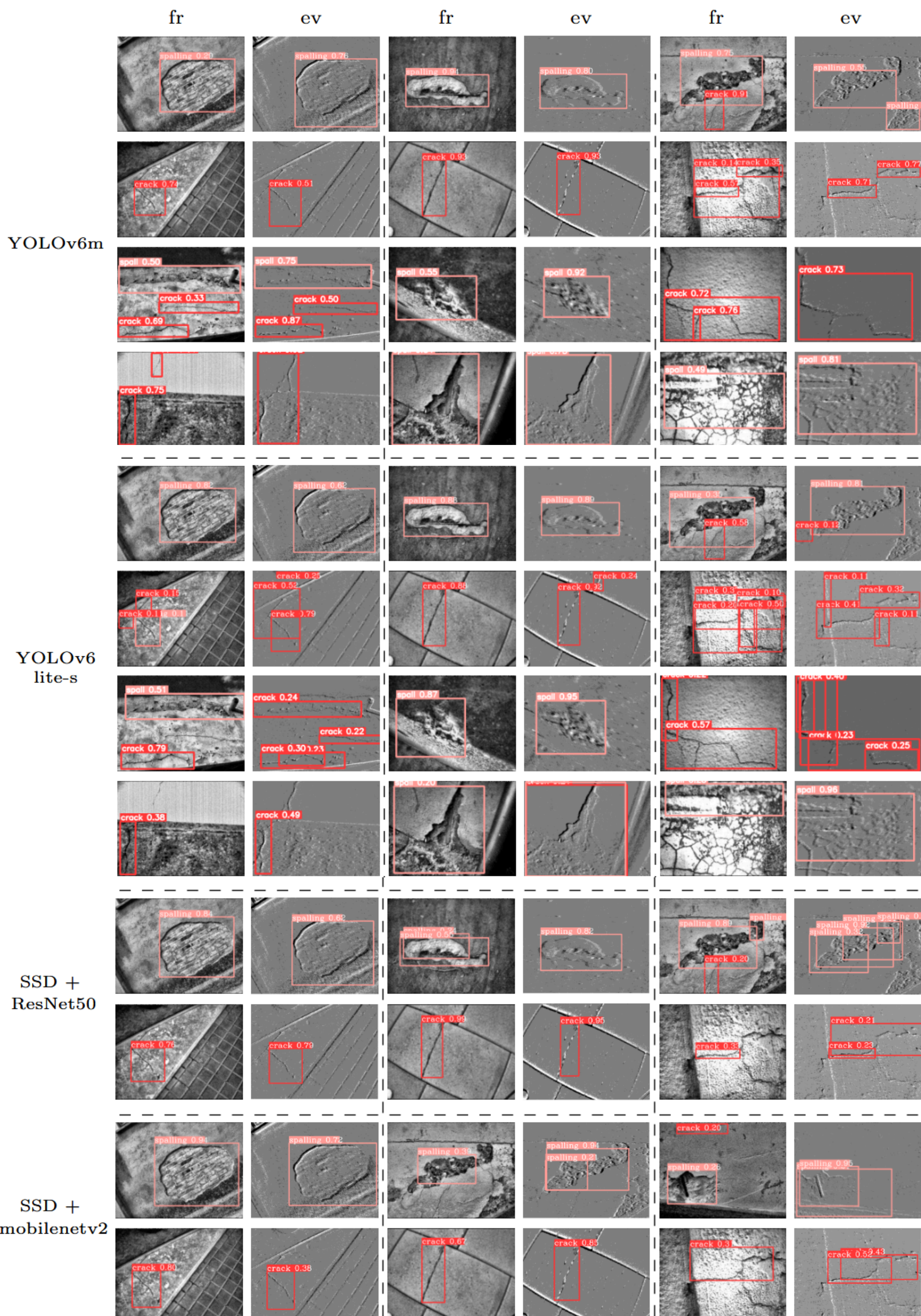

**Figure 14.** Qualititative visualization of event-based and frame-based crack and spalling detection results of four detection models on Adequately-Illuminated Test Set (‘fr’ denotes grayscale frame-based detections, while ‘ev’ represents event-based detections.predicted bounding boxes for cracks are visualized in red color and the predicted bounding boxes for spalling are visualized in pink color)

**Table 6.** Image-based and Event-based performance on Low-Light Test Set

| | $mAP_{0.5:0.95}$ | | | | | | $mAP_{0.5}$ | | | | | | $F1_{IoU0.5}$ | | | | | |
|---|---|---|---|---|---|---|---|---|---|---|---|---|---|---|---|---|---|---|
| | All | | Crack | | Spall | | All | | Crack | | Spall | | All | | Crack | | Spall | |
| | fr | ev | fr | ev | fr | ev | fr | ev | fr | ev | fr | ev | fr | ev | fr | ev | fr | ev |
| YOLOv6m | .04 | **.23** | .09 | **.24** | .01 | **.22** | .13 | **.44** | .21 | **.46** | .04 | **.43** | .25 | **.56** | .33 | **.60** | .15 | **.55** |
| YOLOv6lite-s | .03 | .17 | .05 | .18 | .01 | .15 | .11 | .39 | .16 | .43 | .06 | .35 | .24 | .47 | .28 | .52 | .18 | .45 |
| SSD300-resnet50 | .02 | .17 | .04 | .17 | .01 | .16 | .10 | .38 | .13 | .42 | .07 | .34 | .21 | .47 | .27 | .52 | .15 | .44 |
| SSD300-mobilev2 | .02 | .12 | .04 | .12 | .006 | .13 | .08 | .35 | .11 | .36 | .04 | .34 | .23 | .51 | .27 | .52 | .15 | .51 |

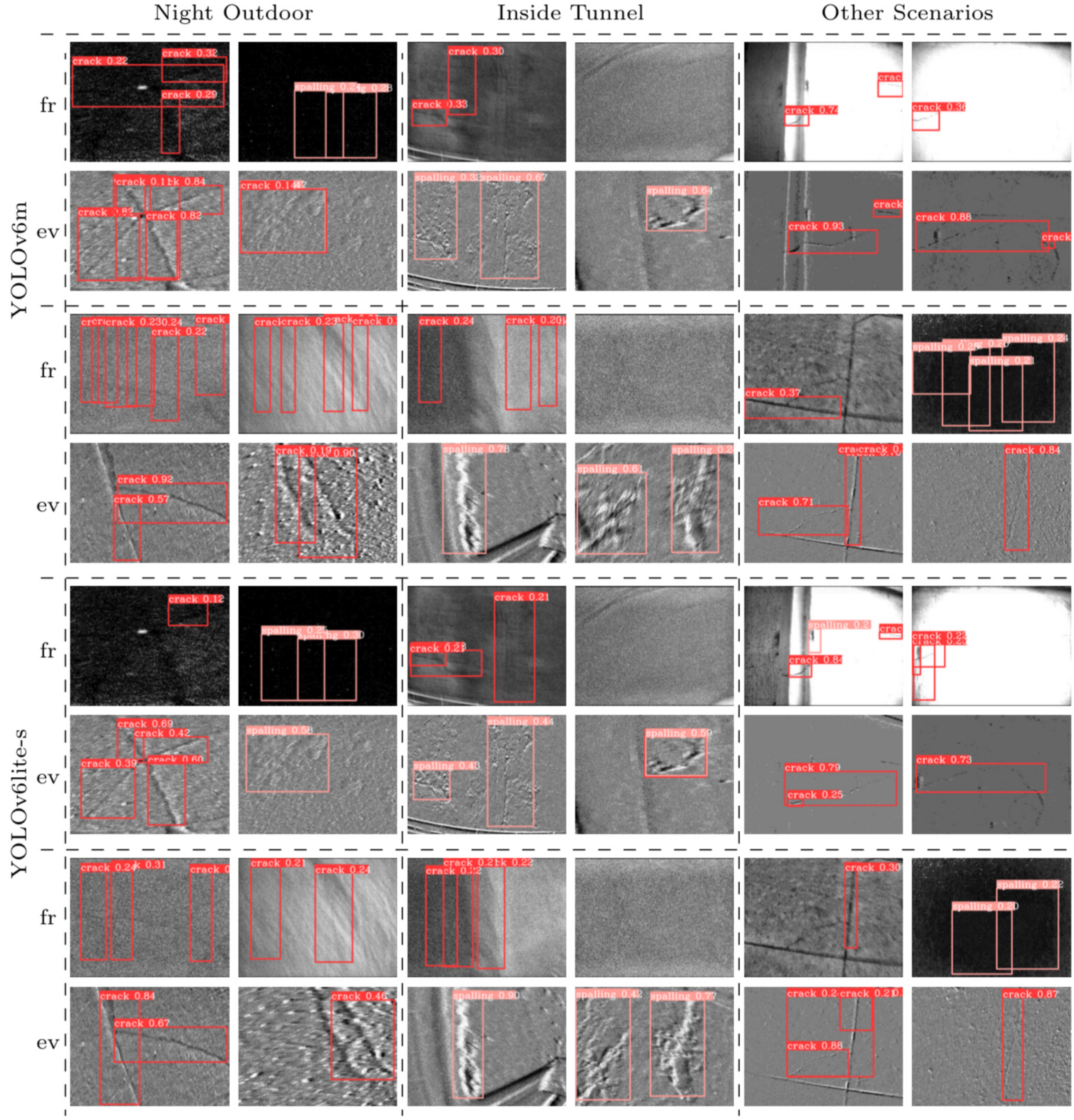

**Figure 15.** Qualitative visualization of the event-based and image-based crack and spalling detection results for the YOLOv6m and YOLOv6lite-s models on the Low-Light Test Set: "Other scenarios" refers to conditions involving saturated and dynamic lighting, in addition to dimly lit environments.('fr' denotes grayscale frame-based detections, while 'ev' represents event-based detections.predicted bounding boxes for cracks are visualized in red color and the predicted bounding boxes for spalling are visualized in pink color)

the training set, thereby aiding in the identification of crack features.

Overall, these results suggest that the laboratory data effectively mitigates the issue of insufficient field data samples in the training set, leading to improved detection

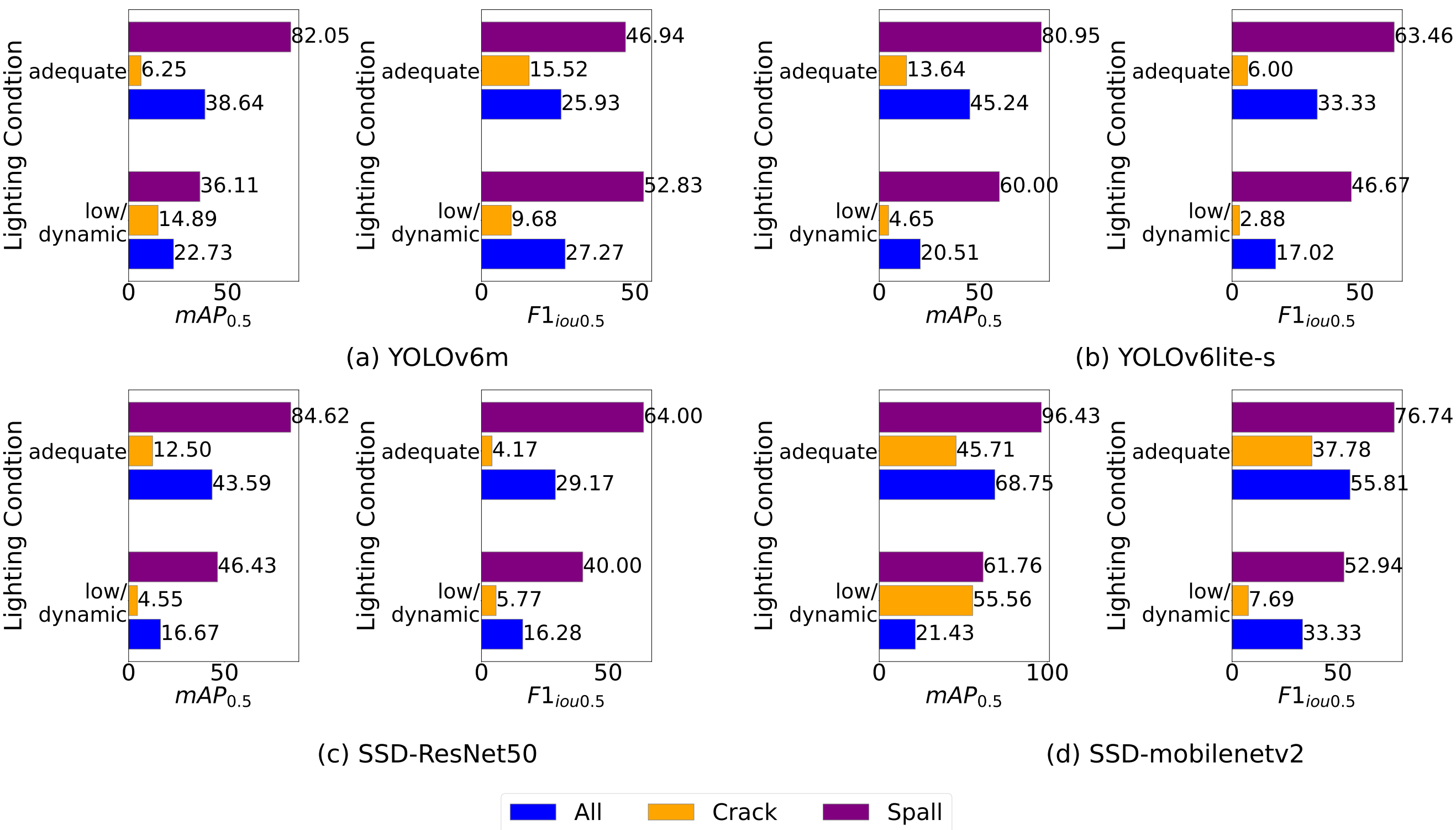

**Figure 16.** Percentage reduction of mAP@0.5 and $F1_{IoU0.5}$ metrics of defect detections with (a) YOLOv6m, (b) YOLOv6lite-s, (c) SSD300 with ResNet backbone, and (d) SSD300 with MobileNetV2 backbone, when those models were trained without Laboratory data. The results are displayed with respect to the two test sets; that is adequate lighting test set and low/dynamic lighting test set

performance on test data comprising exclusively field samples. In other words, even when the detection algorithms are predominantly trained with laboratory data, they remain generalizable and enhance detectability in field data.

### Influence of Temporal Window on 2D Event Histograms for Defect Detection

In this experiment, we evaluated the significance of our proposed adaptive time windowbased 2D histogram formation method, as described in Algorithm 1, in comparison with fixed temporal lengthbased event histogram formation for event-based crack and spalling defect detection.

As shown in subplots (a) and (b) of Fig. 17, compared to event histograms formed using fixed temporal lengths of 10 ms, 15 ms, 20 ms, and 25 ms, our method improves the $mAP_{0.5}$ metric by at least 3.5% and the $F1_{IoU0.5}$ metric by at least 6% for YOLOv6m. For SSD300-ResNet50, the improvements are at least 2% in $mAP_{0.5}$ and 6% in $F1_{IoU0.5}$.

Subplots (c) and (d) in that same Fig. 17 present the detection performance of YOLOv6m and SSD300-ResNet50 on the Low-Light Test Set. In this case, our adaptive temporal windowbased histogram formation method outperforms fixed temporal window approaches of 25 ms, 30 ms, 35 ms, and 40 ms. Here, our method improves $mAP_{0.5}$ by at least 5% for YOLOv6m and 8% for SSD300-ResNet50. The $F1_{IoU0.5}$ metric also shows consistent improvement with our adaptive time window approach, increasing by at least 7% for YOLOv6m and at least 8.5% for SSD300-ResNet50.

### Classification Evaluation

Fig. 18 illustrates the variation in frame-based and event-based classification accuracy of ResNet34, VGG16, EfficientNet-B0, and MobileNetV2 models across Adequately-Illuminated and Low-Light Test Sets, considering four different spatial resolutions of the model inputs($32 \times 32$, $64\times64$, $128\times128$, $224\times224$).

Based on the results, it is evident that both frame-based and event-based models demonstrate nearly equal performance in terms of accuracy when classifying adequately illuminated data. For the Adequately Illuminated dataset, EfficientNet-b0 outperforms all four models, achieving the highest event-based and frame-based classification accuracy of approximately 93% with an input spatial resolution of 224x224.

For the Low-Light Test Set, the highest frame-based classification accuracy achieved was 85%, using the EfficientNet-b0 at an input resolution of $224\times224$, and the MobileNetV2 at input resolutions of both $224\times224$ and $128\times128$. For the same Low-Light Test Set, the highest event-based classification accuracy of 93% was also achieved with the EfficientNet-B0 model at an input resolution of $128\times128$, surpassing the highest frame-based classification accuracy by 8%. Furthermore, the results indicate that, particularly for the event-based classification, using a spatial resolution of $128\times128$ or higher leads to higher accuracy values compared to using lower spatial resolutions of $32\times32$ or $64\times64$.

## Discussion

In this study, we investigated the performance of event-based civil structural defect detection by introducing the ev-CIVIL

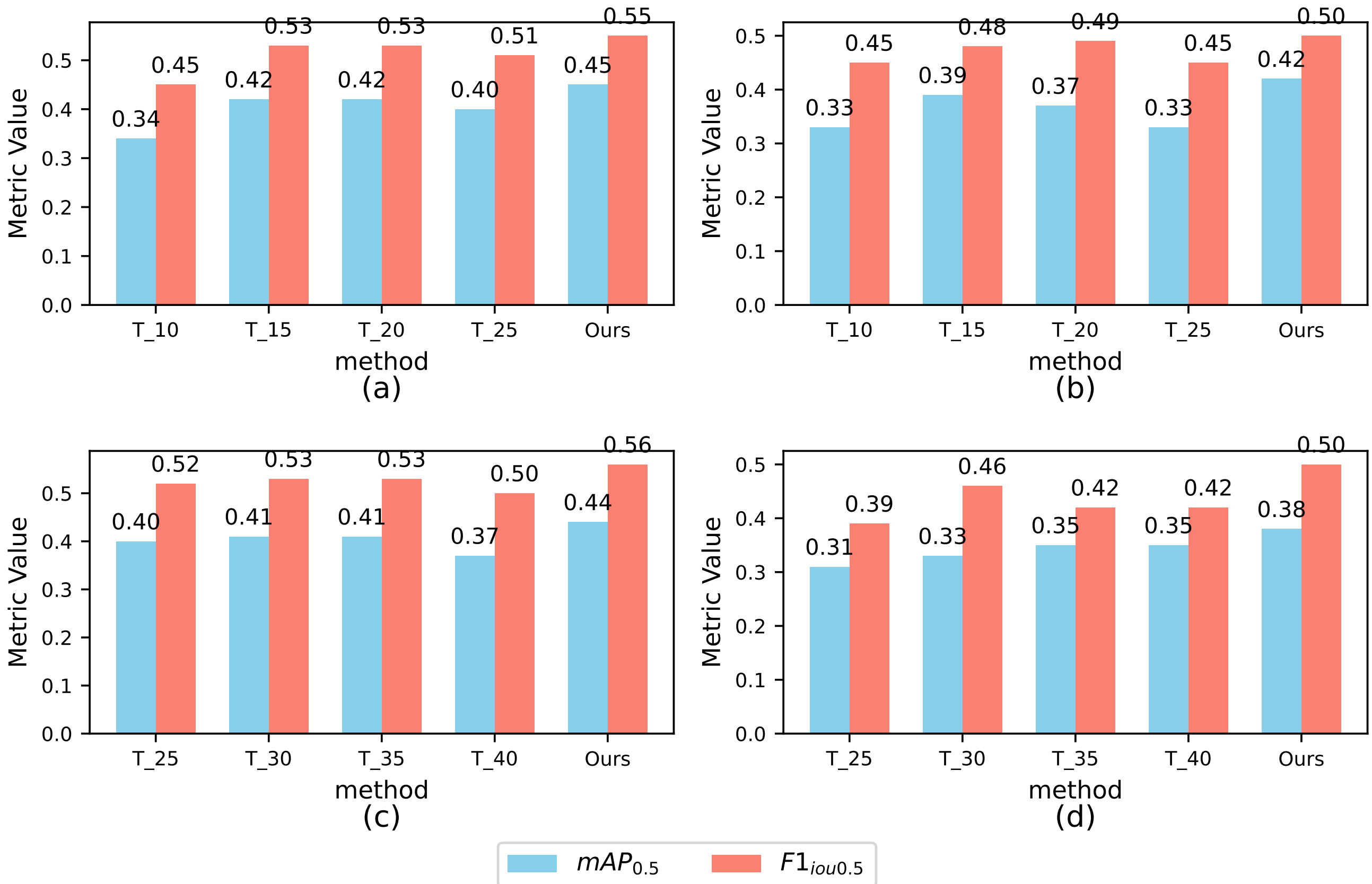

**Figure 17.** Comparison of our 2 channel histogram method explained in Algorithm 1 (Ours) with fixed temporal length based histogram formation method. For adequately illuminated data, the fixed temporal lengths are 10 ms (T_10), 15 ms (T_15), 20 ms (T_20), and 25 ms (T_25). For low-illuminated data, the fixed temporal lengths are 25 ms (T_25), 30 ms (T_30), 35 ms (T_35), and 40 ms (T_40). Performance is evaluated in terms of $mAP : 0.5$ and $F1_{IoU0.5}$.The subplots represent: (a) YOLOv6m performance on the adequately illuminated test set, (b) SSD300-ResNet50 performance on the adequately illuminated test set, (c) YOLOv6m performance on the low-light test set, and (d) SSD300-ResNet50 performance on the low-light test set.

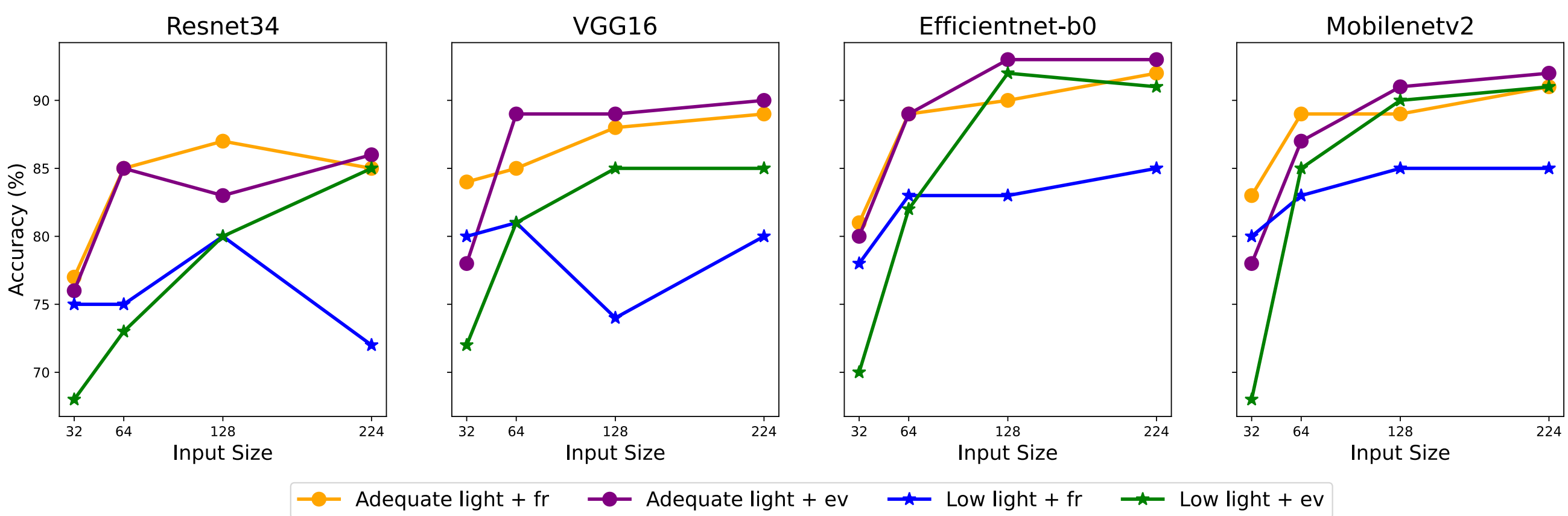

**Figure 18.** Variation in classification accuracy among ResNet34, VGG16, and MobileNetV2 models for frame-based and event-based classification tasks across different input spatial resolutions (32x32, 64x64, 128x128, 224x224) in both Adequately-Illuminated and Low-Light test datasets

dataset. Our research fills a significant gap, as there are currently no datasets or performance analyses specifically for event-based civil structural defect detection. While existing literature on the subject is sparse, we gained insights from related research in event-based object detection, such as vehicle and human detection (referenced as[26],[27],[28] and[30]). Analysis reveals a range of $mAP_{0.5:0.95}$, spanning from 0.18 to 0.5, with $mAP_{0.5}$ falling within the range of 0.25 to 0.75. These variations in performance can be attributed to a variety of factors, including the number of detection classes, the shapes and complexities of the objects, their poses and orientations, sizes and scales, the amount of training data available, and the design and parameters of the detection algorithm. In our experiments using the ev-CIVIL dataset with the YOLOv6m model, we observed $mAP_{0.5:0.95}$ to be 0.22 for nighttime data and 0.24 for daytime data when considering two detection classes. For a single detection class, such as crack detection, YOLOv6m achieved mAP values of 0.27 for nighttime data and 0.30 for daytime data. The relatively lower mAP values observed for crack and spalling defect detection, compared to the typical range of 0.4 to 0.5 reported for car or person detection in the literature, can be attributed to factors such as irregular object shapes and complexities, variability in object scale and orientation, data scarcity, and algorithm sensitivity. In our case, we achieved $mAP_{0.5}$ values of around 0.45 for both nighttime and daytime scenarios with two detection classes, and approximately 0.5 for a single detection class at an IoU threshold of 0.5. These results align well with the range of mAP values reported in the literature for event-based object detection. Additionally, our $mAP_{0.5:0.95}$ also falls within the typical range observed in the literature for this type of detection task. This consistency indicates that our event-based detection approach demonstrates performance levels comparable to those reported in existing research.

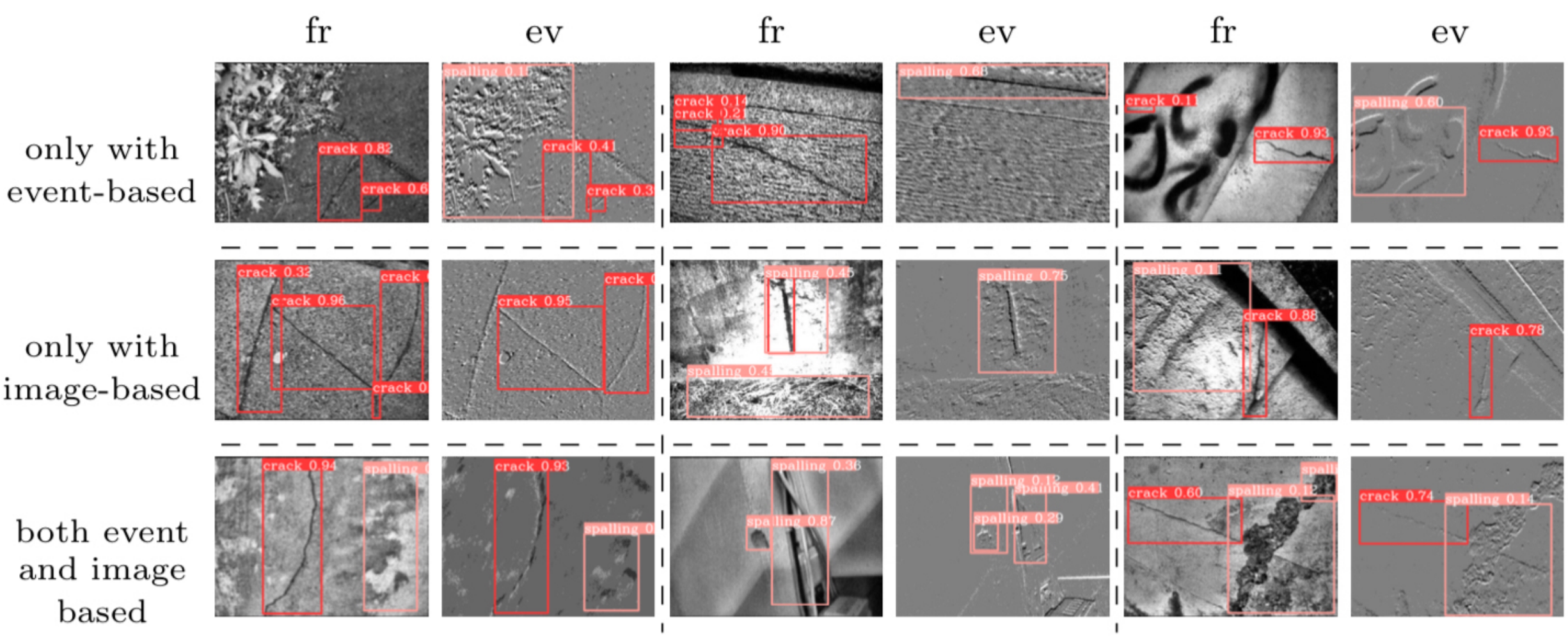

**Figure 19.** Image-based and event-based detection errors ('fr' refers to grayscale frame-based classification, and 'ev' refers to event-based classification)

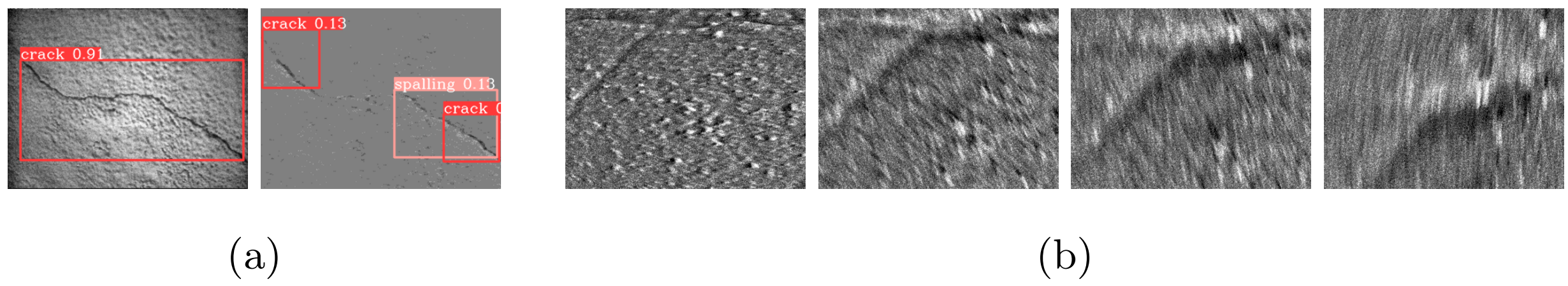

**Figure 20.** Issues involved with event-based detections: (a) partial detections due to directionality of DVS, (b) blur in nighttime event-based 2D histograms as camera movement increases.

This analysis reveals that both detection techniques share numerous similar false positive detections. Fig. 19 shows the error detections (false positives) of the YOLOv6m model for both event-based and frame-based crack and spalling detections. The figure is organized in three rows, each highlighting different aspects of detection errors. The first row illustrates the false positives unique to event-based detections, not found in frame-based detections. This includes instances where bunch of tree leaves and wall arts are sometimes mistakenly identified as spalling, and certain wall edges are misclassified as cracks, exclusively in event-based detections. The second row of the figure highlights the false positives specific to frame-based detections, absent in event-based detections. Noteworthy errors here include misclassification of floor tile edges and exposed iron wires as spalling and the incorrect identification of wall paintings and tree leaves as spalling. The third row showcases cases where false detections occur in both frame-based and event-based detections. Alongside prevIoUsly mentioned error cases, such as detecting wall edges as cracks, this row highlights instances where the removal of wall paintings and a combination of wire and a pipe are erroneously identified as spalling in both detection methods. Based on this analysis, it appears that both image and event-based detections exhibit numerous similar false positive detections, including tree leaves, wall painting removals, and wall edges. This suggests that certain environmental factors or characteristics may contribute to these shared errors across both detection methods In addition to identifying error detections and false positives, we have also observed some limited instances of missed detections in both frame-based and event-based detections. Developing improved object detection models to reduce false positives and miss detections represents a promising direction for future research. This endeavor gains particular significance with the utilization of our newly introduced ev-CIVIL dataset.

In contrast to the advantages of DVSs over traditional frame-based cameras, they exhibit some inherent limitations. One limitation arises from the orientation of defects relative to the camera movement. If the defect orientation aligns with the direction of camera motion, the DVS may emit insufficient events, resulting in high event sparsity and poor defect capture. For example, Fig. 20.a shows a horizontally oriented crack that is not effectively captured because the DVS sensor also moves horizontally.

**Table 7.** Comparison of frame-based (fr) and event-based (ev) inference rates (fps) on an NVIDIA RTX 3090 GPU (24 GB VRAM) and an Intel Core i9 14th-generation CPU (fps : frames per second , 'B' refers to the batch size)

| | GPU | | | | CPU | |
|---|---|---|---|---|---|---|
| | B = 1 | | B = 16 | | fr | ev |
| | fr | ev | fr | ev | fr | ev |
| YOLOV6m | 169 | 116 | 243 | 161 | 8 | 7 |
| YOLOv6lite-s | 206 | 133 | 478 | 239 | 43 | 38 |
| SSD300-resnet50 | 318 | 285 | 460 | 434 | 21 | 19 |
| SSD300-mobilev2 | 347 | 264 | 512 | 333 | 54 | 50 |

Another limitation is that DVSs produce a higher number of noisy spikes and increased motion blur, particularly under nighttime conditions[33], which reduces the accuracy and robustness of defect detection. As shown in Fig. 20.b, event-based 2D histograms become increasingly blurred as camera movement speed increases (left to right) under low-light conditions.

To mitigate these limitations, combining DVS data with grayscale or RGB images or depth point clouds can be an effective strategy. For instance, when inspecting poorly lit tunnels, DVS sensors can overcome low-light issues, while the simultaneous acquisition of image frames can compensate for DVS noise. Although the image frames may be of lower quality, fusing the two modalities using an appropriate deep learning model can enhance detection performance, outperforming single-modality approaches. Hybrid sensing approaches have shown promise in robotics and navigation applications[77] and may open new avenues for UAV-based civil infrastructure inspection. In these inspections, high-resolution DVS cameras[55] allow drones to maintain a safe distance from the inspection surfaces while still capturing sufficient event information.

Another drawback of DVSs is the additional computational burden required to convert asynchronous event streams into usable formats, such as 2D two-channel event histograms. Table 7 compares the inference frame rates of grayscale frame-based detection and event-based detection on both GPU and CPU for the benchmarking models. While frame-based detection generally achieves higher frame rates, both methods remain real-time on GPU. Except for YOLOv6m model, all the other models can be considered as real-time ($fps > 15$) on CPU also for both event-based and frame-based inference.

The ev-CIVIL dataset offers multiple opportunities for refinement, including the implementation of defect segmentation, broadening the variety of defect types, augmenting the dataset with more instances of existing defects, capturing data using a high-resolution DVS camera, and exploring data acquisition via small UAV-based methods. These enhancements represent promising avenues for future research and development, building upon the foundation established by our current study.

## Conclusion

This study explores the use of DVS sensors or event-based cameras for civil infrastructure defect detection, introducing the ev-CIVIL dataset, which includes data from field and laboratory environments under various lighting conditions. We benchmarked four object detection models (YOLOv6m, YOLOv6lite-s, SSD300 with ResNet50, and SSD300 with MobileNetV2) and four classification models (ResNet34, VGG16, EfficientNet-b0, and MobileNetV2). Event-based detection outperformed frame-based methods in low-light conditions, with YOLOv6m showing the highest defect detection performance. Notably, event-based detection surpassed frame-based detection by 0.20 to 0.30 in the Low-Light Test Set. The laboratory dataset was crucial for training models to detect spalling defects, as field data with spalling defects was limited and most available field data was used for the test set. In classification tasks, EfficientNet-b0 achieved the highest accuracy, with event-based models performing better in low-light scenarios (92% vs. 85% for frame-based models). This study also identifies key limitations in data collection and event-based detection methodologies, while highlighting opportunities for future research to improve and expand upon these findings.

## Acknowledgements

This project has received funding from the European Unions Horizon 2020 research and innovation programme under the Marie Sklodowska-Curie grant agreement No: 953454.

# APPENDIX

## YOLOv6 models

In our evaluations, we used the YOLOv6m[16] and YOLOv6-lite[16] variants of the YOLOv6 architecure. The YOLOv6 architecture consists of three major components: the backbone network, the neck, and the efficient decoupled head, as depicted in Fig. 22, with respect to the specific components of the YOLOv6m model.

The backbone network in any object detection architecture primarily serves the purpose of feature extraction. For YOLOv6m, the backbone uses EfficientRep[65] architecture, which incorporates advanced architectural features like RepVGG[64]. This allows the model to train with more complex structures and deploy a simplified version, striking a balance between lightweight design and high accuracy. The backbone also uses a CSP[66]-like design to improve gradient flow and feature reuse while keeping computational costs manageable. These optimizations enable YOLOv6m to be both accurate and a real-time detector, making it suitable for demanding tasks where both speed and precision are required, such as video surveillance or autonomous driving. The EfficientRep backbone focuses on extracting high-quality features with fewer operations compared to conventional backbones like ResNet[19] and CSPDarknet[66], ensuring real-time performance without compromising on accuracy. In contrast, YOLOv6-lite employs a more lightweight backbone, optimized for ultra-fast inference on resource-constrained devices. The YOLOv6-lite backbone uses fewer layers and channels compared to YOLOv6m, ensuring faster processing and reduced computational overhead. It integrates efficient architectures such as GhostNet[67] and ShuffleNet[68], which leverage depthwise separable convolutions and group convolutions to reduce the number of parameters and operations. This design results in a model that is exceptionally lightweight, fast, and well-suited for real-time applications on edge devices with limited resources (e.g., memory and computational power), while still maintaining sufficient accuracy for a wide range of object detection tasks.

The neck in object detection architectures plays a vital role in fusing and refining multi-scale features extracted from the backbone network. In YOLOv6m, the neck incorporates a combination of Feature Pyramid Network (FPN)[69] and Path Aggregation Network (PAN)[70]structures. The FPN captures rich semantic information across different scales, while the PAN enhances the flow of spatial and low-level details, which are crucial for precise object localization. Together, these structures enable YOLOv6m to excel at detecting both small and large objects, ensuring high accuracy and robust performance. In contrast, the YOLOv6-lite neck adopts a simpler design tailored for fast processing on devices with limited power and memory. It simplifies the FPN and PAN structures by reducing the number of layers, channels, and overall complexity. Additionally, it integrates efficient operations such as depthwise separable convolutions and bottleneck structures, minimizing computational costs while preserving essential features for detection.

The head of an object detection model is responsible for generating the final predictions, including bounding boxes, class probabilities, and confidence scores. Both YOLOv6m and YOLOv6-lite employ decoupled heads, which separate the bounding box regression task from the classification task. This decoupled design improves performance by allowing each task to specialize independently, ultimately enhancing both accuracy and efficiency. In YOLOv6m, the decoupled head is more complex, with a larger number of parameters and channels, enabling more precise and robust predictions. On the other hand, YOLOv6-lite uses a streamlined version of this decoupled design, reducing its computational complexity and parameter count. This lightweight approach is optimized for real-time performance on resource-constrained devices, trading off some accuracy for speed and efficiency. Additionally, both models utilize an anchor-free detection strategy, meaning that they do not rely on predefined anchor boxes for object localization. Instead, YOLOv6 directly predicts the bounding box coordinates relative to object centers, making the detection process more flexible and efficient. This anchor-free approach further contributes to faster inference and better adaptability to various object shapes and sizes, particularly in real-time applications where computational efficiency is crucial.

## SSD models

:

In SSD300-ResNet50, as shown in Fig. 21 the backbone uses ResNet50[19], a deep residual network known for its ability to capture high-level, semantic features. ResNet50 generates feature maps at different layers, which are then utilized by the SSD head which consists of additional feature layers designed to detect objects across multiple spatial resolutions. These feature maps correspond to objects of varying sizes, enabling the network to perform well on both small and large objects. However, the deeper architecture of ResNet50 results in higher computational costs and memory usage, which can limit its applicability for real-time applications on devices with limited computational resources. In SSD300-MobileNetV2, the backbone employs the MobileNetV2[49] architecture, designed for lightweight and efficient computation. MobileNetV2 uses depthwise separable convolutions, which significantly reduce the number of parameters and operations compared to traditional convolutions. This results in faster processing and greater efficiency, making SSD300-MobileNetV2 ideal for resource-constrained environments. However, this comes with a slight trade-off in feature quality, as the model sacrifices some level of detail in favor of speed and efficiency.

The head of an SSD architecture is responsible for generating predictions, including bounding boxes, class labels, and confidence scores. SSD uses default boxes (also known as anchor boxes), which are predefined bounding boxes with different aspect ratios and scales. These anchors are placed at various locations on the multi-scale feature maps generated by the backbone, enabling the model to detect objects of different sizes. In both SSD300-ResNet50 and SSD300-MobileNetV2, the head processes the feature maps from multiple stages of the backbone, using 8732 default boxes across different spatial resolutions. Each default box is associated with anchors that predict the bounding box coordinates and object classification scores. However, due to the lightweight design of the MobileNetV2 backbone, the feature maps are less detailed compared to those from ResNet50, leading to slightly reduced accuracy, especially for detecting smaller objects. Despite this, the SSD300-MobileNetV2 head still performs multi-scale detection using efficient anchor matching, enabling faster inference with reduced computational and memory requirements making it suitable for real-time processing on edge devices.

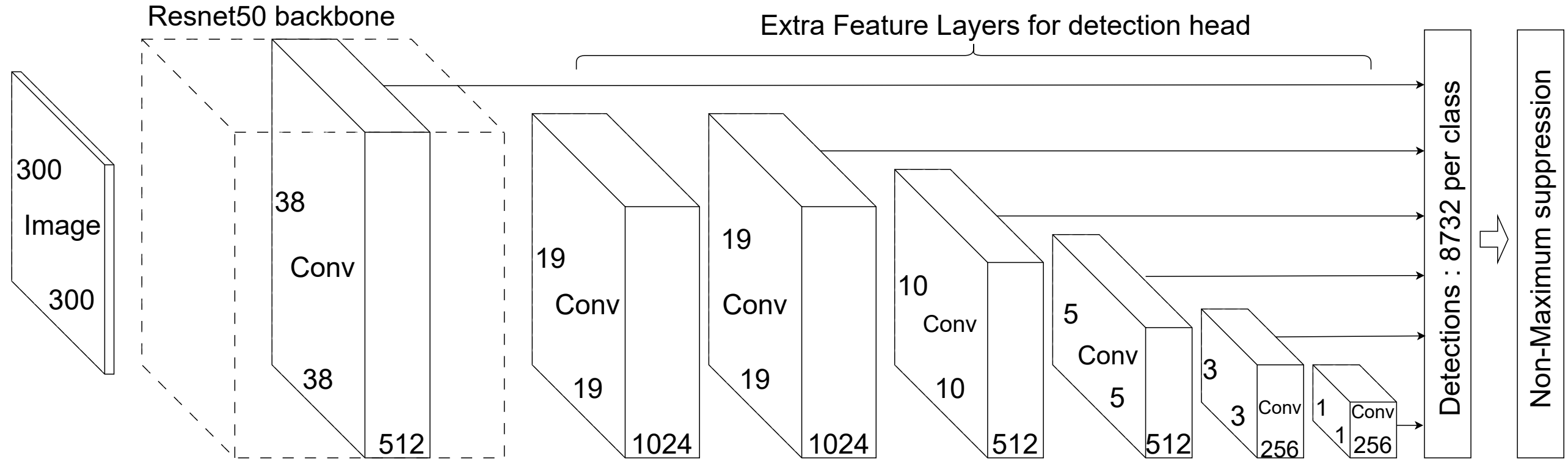

**Figure 21.** SSD300 Architecture

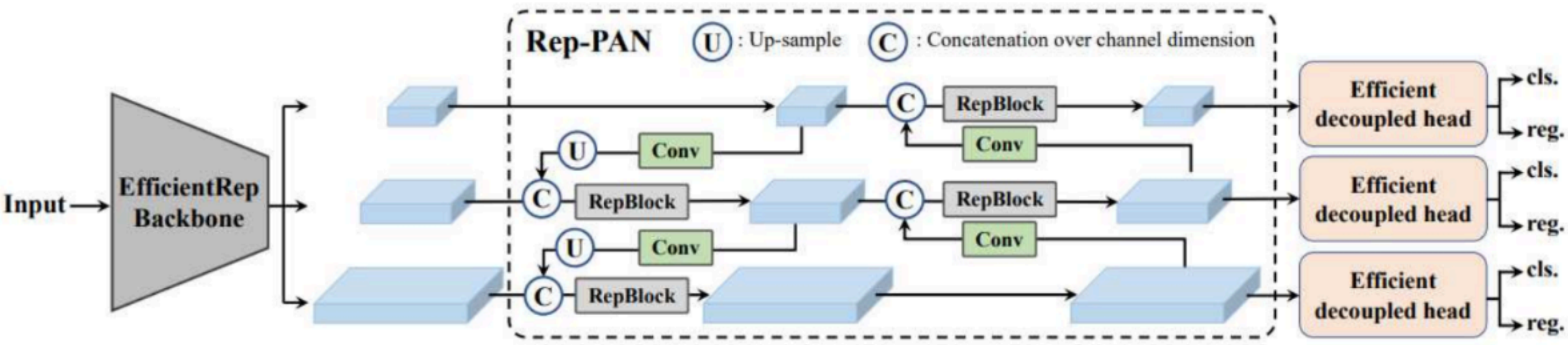

**Figure 22.** Yolov6 Architecture[71]

## *Evaluation Metrics explanation*

*mean average precision (mAP)* :

The mAP is computed as the average of AP values across all classes $c$ that needs to be detected(in our case crack and spalling) as in Eq. (3).

$$mAP = \frac{1}{C}\sum_{c=1}^{C} AP(c) \tag{3}$$

where,

C : total number of classes

$AP(c)$ : is calculated as in Eq. (4)

For any class $c$ that needs to be detected, the average Precision (AP) over different IoU thresholds can be calculated as in Eq. (4)

$$AP(c) = \frac{1}{n}\sum_{i=1}^{n} P(IoU_i) \tag{4}$$

where,

$n$ : the number of IoU thresholds. (More specifically, for COCO $AP_{0.5:0.95}$, these IoU thresholds range from 0.5 to 0.95 with a step size of 0.05. And for COCO $AP_{0.5}$ n = 1, as it calculated for the IoU threshold 0.5)

$P(IoU_i)$ : the precision calculated at IoU threshold $IoU_i$ as in Eq. (6)

$F1_{IoU0.5}$ *score* :

$F1_{IoU0.5}$ metric which combines precision and recall to provide a balanced measure of the model's ability to detect objects accurately and completely, particularly at an IoU threshold of 0.5. So in calculating the F1 score at IoU 0.5 as shown in equation Eq. (5), it means that only the detections with an IoU greater than or equal to 0.5 with the ground truth are considered as correct detections. For Eq. (5) the $P(IoU_{0.5})$ and $R(IoU_{0.5})$ is calculated using Eq. (6) and Eq. (7) with $i = 0.5$.

$$F1_{IoU0.5} = \frac{2 \times P(IoU_{0.5}) \times R(IoU_{0.5})}{P(IoU_{0.5}) + R(IoU_{0.5})} \tag{5}$$

The precision (P) at a specific Intersection over the Union (IoU) threshold is calculated as Eq. (6)

$$P(IoU_i) = \frac{TP(IoU_i)}{TP(IoU_i) + FP(IoU_i)} \tag{6}$$

where,

$TP(IoU_i)$ : Number of true positive detections at the IoU threshold $IoU_i$

$FP(IoU_i)$ : Number of false positive detections at the IoU threshold $IoU_i$

The Recall (R) at a specific IoU threshold $IoU_i$ is calculated as Eq. (7)

$$R(IoU_i) = \frac{TP(IoU_i)}{TP(IoU_i) + FN(IoU_i)} \tag{7}$$

where,

$TP(IoU_i)$ : Number of true positive detections at the IoU threshold $IoU_i$

$FN(IoU_i)$ : Number of false negative detections at the IoU threshold $IoU_i$